\documentclass{article} 
\usepackage[submission]{colm2025_conference}

\usepackage{microtype}
\usepackage{hyperref}
\usepackage{url}
\usepackage{booktabs}
\usepackage{amsmath}
\usepackage{multirow} 
\usepackage{lineno}
\usepackage{enumitem}
\usepackage{graphicx}
\usepackage{subcaption}
\definecolor{darkblue}{rgb}{0, 0, 0.5}
\hypersetup{colorlinks=true, citecolor=darkblue, linkcolor=darkblue, urlcolor=darkblue}

\title{On the transferability of Sparse Autoencoders for interpreting compressed models}

\author{Suchit Gupte, Vishnu Kabir Chhabra \& Mohammad Mahdi Khalili \\
Department of Computer Science and Engineering\\
The Ohio State University\\
Columbus, OH 43210, USA \\
\texttt{\{gupte.31,chhabra.67,khalili.17\}@osu.edu} \\
}

\begin{document}

\maketitle

\begin{abstract}
Modern LLMs face inference efficiency challenges due to their scale. To address this, many compression methods have been proposed, such as pruning and quantization. However, the effect of compression on a model's interpretability remains elusive. While several model interpretation approaches exist, such as circuit discovery, Sparse Autoencoders (SAEs) have proven particularly effective in decomposing a model's activation space into its feature basis. In this work, we explore the differences in SAEs for the original and compressed models. We find that SAEs trained on the original model can interpret the compressed model albeit with slight performance degradation compared to the trained SAE on the compressed model. Furthermore, simply pruning the original SAE itself achieves performance comparable to training a new SAE on the pruned model. This finding enables us to mitigate the extensive training costs of SAEs. For reproducibility, we make our code available at \url{https://github.com/osu-srml/sae-compression.git}

\end{abstract}

\section{Introduction}
Modern transformer-based language models \citep{Vaswani+2017,guo2025deepseek,team2024gemma,touvron2023llama} record high performance, however, suffer in the realm of efficiency due to their scale. To address this concern the field of efficient neural network inference, model compression, has garnered attention \citep{zhou2024survey,xiao2023smoothquant,shao2023omniquant,lin2024awq,bai2022towards}.
The most popular methods of compression have shown to be quantization \citep{lin2024awq,shao2023omniquant,xiao2023smoothquant}, pruning \citep{wanda,frantar2023sparsegpt,ma2023llm} and low-rank decomposition \citep{saha2024compressinglargelanguagemodels,hsu2022languagemodelcompressionweighted}. While research in the domain has found success in making large language models more efficient for inference, concerns regarding safety of models that underwent compression remain a persistent issue \citep{hong2024decoding,xu2024perplexitymultidimensionalsafetyevaluation}.  
In conjunction, mechanistic interpretability has garnered attention. Research in the field aims to elucidate the underlying mechanisms of language models using methods such as circuit analysis \citep{wang2022interpretability,prakash2024fine,hanna2023does,conmy2023towards,chhabra2025neuroplasticity}, direction feature attribution \citep{arditi2024refusal,makelov2024principledevaluationssparseautoencoders} etc. 
Recent works in mechanistic interpretability aim to aid the safety of language models via interpreting and identifying features present in language models utilizing meta-models called Sparse Autoencoders (SAEs) \citep{cunningham2023sparse}. Sparse Autoencoders aim to decompose features from superposition, essentially providing a useful decomposition of the activation space \citep{cunningham2023sparse, lan2024sparse,paulo2025sparseautoencoderstraineddata}. However, training SAEs incur significant computational costs \citep{gao2024scaling}.

While work in mechanistic interpretability has aimed to elucidate complex mechanism via the use of sparse autoeconders , circuit analyis etc. Research aiming to better understand the utility and interpretability such techniques for model that underwent compression is limited. Our work focuses on aiming to better understand the use of sparse autoencoders for compressed models.

In our work we aim to improve the interpretability and safety of compressed models. We specifically aim to utilize SAEs in the pruned model and explore the relations of the SAE of the original model and the SAE on the pruned model. Our findings can be summarized as follows: 
\begin{itemize}
    \item We show that the SAE of the original model can be transferred to the pruned model and is comparable to the SAE trained on the pruned model. 
    \item We further show that utilizing the pruning method on the SAE itself can lead to an SAE that performs as well as the SAE trained on the pruned model. 
\end{itemize}
\section{Preliminaries}
This section outlines the key concepts that form the foundation of our work, including sparse autoencoders (SAEs), neural network pruning methods, and the \textit{WANDA} pruning approach used to compress large language models (LLMs).
\subsection{Sparse autoencoders}
Sparse autoencoders (SAEs) have emerged as an unsupervised technique for interpreting the internal activations of large language models (LLMs). By representing activation vectors as sparse linear combinations of learned directions, SAEs can uncover potentially interpretable structure within the model’s representation space (\cite{bricken2023interpreting, cunningham2023sparse}). The core idea is to encode each activation into a sparse latent representation and reconstruct the original input using a learned dictionary of directions.

An SAE is a linear encoder-decoder neural architecture. The encoder transforms an input activation $x \in \mathbb{R}^n$ into a sparse higher-dimensional latent vector $h \in \mathbb{R}^M$, where $M \gg n$, while the decoder reconstructs the input using this sparse representation:
\begin{align}
    \text{\textbf{Encoding step:}} \hspace{10px} h &= \sigma(W_E x + b_E) \\
    \text{\textbf{Decoding step:}} \hspace{10px} \hat{x} &= W_D h + b_D
\end{align}

$W_E, b_E$ are the encoder weights and biases, and $W_D, b_D$ are the decoder weights and biases. $\sigma$ is the activation function. Training this neural architecture involves minimizing a loss function that combines reconstruction error with a sparsity-inducing $L_1$ penalty:
\begin{align}
    \text{\textbf{Loss function:}} \hspace{10px} \mathcal{L} = \|x - \hat{x}\|_2^2 + \lambda \|h\|_1,
\end{align}

The hyperparameter $\lambda$ acts as a regularizer controlling the sparsity level in the learned representation. The $L_1$ penalty on $h$ encourages the model to reconstruct inputs using a sparse linear combination of the latents. This approach has been validated (\cite{leeshark, wright}) as an effective way to recover the ground-truth latents. 

The choice of activation function plays a critical role in enforcing sparsity. Early works relied on \textit{ReLU} activation for its simplicity and non-negativity constraint. More recent works have explored other activation functions like \textit{TopK} (\cite{topk}), \textit{BatchTopK} (\cite{batchtopk}), \textit{JumpReLU} (\cite{jumprelu}), and \textit{ProLU} (\cite{prolu}). 

\subsection{Pruning techniques}
Pruning is a widely used technique to reduce the size and computational footprint of neural networks by eliminating redundant or low-impact weights, while preserving model performance. In the context of large language models (LLMs), pruning becomes particularly important due to their massive scale and high inference costs.

A common baseline is \textit{Magnitude} pruning (\cite{magnitude}), which ranks and removes weights with the smallest absolute values under the assumption that such weights contribute less to the model's predictions. Although simple and effective for smaller networks, this method performs poorly on large-scale models. \textit{SparseGPT} (\cite{sparsegpt}), a more recent approach, leverages layerwise weight reconstruction involving inverse Hessian computations. Although accurate, this approach is computationally intensive and requires weight updates, making them impractical at scale. 

\textbf{WANDA} (Pruning by \textbf{W}eights \textbf{and A}ctivations, \cite{wanda}) offers a fast and effective strategy for pruning pretrained models without any retraining or gradient-based updates. \textit{WANDA} extends \textit{Magnitude pruning} by incorporating input activations, producing a more faithful estimate of weight importance. The importance score for each weight is defined as:
\begin{align}
    S_{ij}^{\text{Magnitude}}=|W_{ij}| \hspace{10px} \longrightarrow \hspace{10px} S_{ij}^{\text{WANDA}}=|W_{ij}|.||X_j||_2 
\end{align}

where $||X_j||_2$ is the $L_2$ norm of the $j^{th}$ input activation across the token batch. 

\textit{WANDA} offers a simple yet highly effective approach to pruning large language models. By combining weight magnitudes with input activation strength, it captures which weights are both large and frequently used, leading to more informed pruning decisions. The method operates row-wise, ensuring balanced sparsity across output neurons, and requires no retraining, no gradient updates, and only a single forward pass over a small calibration set. Despite its simplicity, \textit{WANDA} demonstrates performance comparable to more compute-heavy approaches like \textit{SparseGPT} on zero-shot and language modeling tasks while being significantly more efficient and scalable (\cite{wanda}).
\section{Experimental Setup}
To investigate the interplay between pruning and SAE training, we propose the following hypothesis,
\begin{center}
\framebox{%
\parbox{0.95\textwidth}{\textbf{Pruning an SAE originally trained on the full, unpruned LLM results in a new SAE that exhibits similar behavior to an SAE trained on a \textit{WANDA}-pruned LLM.} }}
\end{center}

The following experimental setup is designed to validate the above hypothesis, enabling efficient post hoc analysis without the computational overhead of training SAEs on the pruned model if the pretrained SAEs exist.
\subsection{Model Configurations}\label{Exp setup: Model config}
\textbf{Architectures:} Our experiments focus on two transformer models: GPT-2 \citep{gpt2} and Gemma-2-2B \citep{team2024gemma}. GPT-2 is a 12-layer decoder-only transformer, while Gemma-2-2B is a larger model with 26 decoder-only layers. Both architectures follow the standard transformer design with multi-head self-attention followed by a feed-forward MLP sublayer, with residual connections and normalization.

To extract meaningful intermediate representations for SAE training, we target three key activation sites in each transformer block: the attention output (accessed via \texttt{attn.hook\_z}), the MLP output (\texttt{hook\_mlp\_out}), and the post-MLP residual stream (\texttt{hook\_resid\_post}). These hooks correspond to those implemented in the \textit{transformer\_lens} library \citep{nanda2022transformerlens}. 


\textbf{Pruning with \textit{WANDA}}: We apply \textit{WANDA} pruning \citep{wanda} with 50\% sparsity to both GPT-2 and Gemma-2-2B.\footnote{In this paper, when we say $x \%$ sparsity rate, it implies that $x \%$ of the model parameters are pruned. } Pruning is applied to the query, key, value, and output projection matrices of the attention mechanism ($\textbf{W}_\textbf{Q}$, $\textbf{W}_\textbf{K}$, $\textbf{W}_\textbf{V}$, and $\textbf{W}_\textbf{O}$), as well as the input and output projections of the MLP block ($\textbf{W}_\textbf{in}$ and $\textbf{W}_\textbf{out}$).

For pruning calibration, we use 128 context-length sequences sampled from the OpenWebText dataset \citep{Gokaslan2019OpenWeb}), following the pruning configurations defined in the original \textit{WANDA} paper.  All evaluations are conducted using these pruned models to maintain consistency across SAE variants.

\subsection{SAE Configurations}\label{Exp setup: Sae config}
\textbf{Pretrained SAEs:} We utilize existing pretrained SAEs for both models. For GPT-2, we use SAEs from the \textit{sae\_lens} library (\cite{bloom2024saetrainingcodebase}, which were trained on OpenWebText activations. For Gemma-2-2B, we use SAEs provided by the Gemma Scope work \citep{lieberum2024gemmascope}), trained on the Pile dataset \citep{gao2020pile800gbdatasetdiverse}). These SAEs operate on normalized activations, use the \textit{JumpReLU} activation function (\cite{jumprelu}), and reconstruct inputs using overcomplete dictionaries of latent directions.

\textbf{SAE Training on Pruned Models:} To evaluate the effects of pruning on SAE training, we train SAEs from scratch using activations extracted from the pruned GPT-2 and Gemma-2-2B models. For each model, we target the same three activation sites used during pretrained SAE training. The training datasets (OpenWebText for GPT-2 and the Pile for Gemma-2-2B) are held constant to ensure fair comparisons. Our training closely follows the respective pretrained SAE configurations.


\textbf{Pruning of Pretrained SAEs:} We also simulate pruning-aware SAE training by directly pruning the encoder and decoder weights of the pretrained SAEs ($\textbf{W}_\textbf{E}$ and $\textbf{W}_\textbf{D}$). Using the same datasets employed during SAE training, we sweep across a range of sparsity levels and select the level (at least 25\%) at which the pruned SAE achieves reconstruction loss comparable to its trained counterpart (the one trained on the pruned model).

\subsection{Evaluation Metrics} \label{Exp setup: Eval metric config}
To holistically validate our hypothesis, we use \textit{SAEBench} \citep{SAEBENCH}, a comprehensive evaluation suite designed to rigorously assess the performance and quality of SAEs. \textit{SAEBench} goes beyond conventional evaluation by probing four key dimensions of SAEs: concept detection, interpretability, reconstruction fidelity, and feature disentanglement. Together, these dimensions assess whether individual features are semantic, human-interpretable, retain model behavior, and distinct. Using this framework, we rigorously evaluate a pruned SAE to ensure that its learned representations remain  functionally aligned with those of the corresponding trained SAE. Please see Appendix \ref{Section: Appendix eval metric cfg} for more details on the setup of each metric.
\subsubsection{Unsupervised Metrics (Core)}
The Core metrics include standard measures such as sparsity and reconstruction quality. The reconstruction quality is assessed using Mean Squared Error, explained variance to capture the proportion of input variance retained, and cosine similarity to measure the directional alignment between original and reconstructed activations. 

We evaluate the influence of SAEs on model behavior by analyzing shifts in output logits using KL divergence and a normalized cross-entropy loss score. This score compares the original loss to the loss after replacing layer activations with SAE reconstructions, where values closer to 1 reflect better preservation of the model’s predictive behavior.

We measure sparsity using $L_0$ sparsity to capture the average number of non-zero activations and $L_1$ sparsity to reflect total activation strength. We track feature density to identify dead features that never activate and overly dense features that activate too often, both of which reduce interpretability. While these metrics offer valuable insights into SAEs' representational and functional behavior, they are insufficient for evaluating SAEs.

\subsubsection{Feature Absorption}
Feature absorption is a sparsity-driven phenomenon in SAEs where, given hierarchical concepts where \textit{A} implies \textit{B} (e.g., India implies Asia), the model favors learning a latent for \textit{A} and another for \textit{B except A} to reduce redundancy. While this encourages sparsity, limiting each input to a single active latent can hurt interpretability by fragmenting the representation of concept \textit{B}. As a result, features can behave unpredictably; for instance, a latent meant to detect if a word starts with \textit{S} might work in most cases but randomly fail in some because those cases have been \textit{absorbed} by other features.

To quantify this, \cite{SAEBENCH} extend the work by \cite{ABSORPTION} using a first-letter classification task with logistic regression probes to identify main SAE latents. Test tokens are analyzed where primary latents fail, yet the probe is able to correctly classify. If another latent activates on these test tokens and strongly aligns with the probe direction through cosine similarity and projection contribution (\cite{SAEBENCH_blog}), it is identified as an absorbed feature. In contrast to \cite{ABSORPTION}, which relied on ablation to identify absorption, \cite{SAEBENCH} directly measure each latent’s contribution to the probe, providing a more consistent and reliable evaluation across layers.

\subsubsection{SCR and TPP}
Spurious Correlation Removal (SCR) helps to remove hidden biases in a model’s behavior. Models often pick up spurious patterns, such as linking gender with profession, because these correlations appear in the training data. \cite{SAEBENCH} extend the work by \cite{SHIFT}, where the idea is to train a binary classifier that has learned the target variable and misleading correlations. The idea is to determine which latents in the SAE are responsible for the misleading correlations. By zeroing out a small set of these latents, \cite{SAEBENCH} measure the reduction in spurious influence. If these latents are truly tied to the spurious traits, their removal improves the classifier’s focus on the intended target.

Targeted Probe Perturbation (TPP) extends SCR to multi-class scenarios. It trains binary probes for each individual class to identify the most important latents associated with each class. The impact of zero-ablating these latents is then evaluated by observing changes in probe accuracy across all classes. A high TPP score suggests that different concepts are represented by separate groups of latents. Ablating the latents tied to one class should mainly reduce probe accuracy for that class while having a minimal effect on other probes.

\subsubsection{RAVEL}
An effective SAE encodes each concept in separate latents, achieving clear disentanglement. RAVEL (Resolving Attribute–Value Entanglements in Language Models) tests whether an SAE encodes different concepts in separate latents. RAVEL tests if a single attribute can be modified independently by tweaking its corresponding latent value in the SAE and checking whether only that attribute changes in the model’s output. RAVEL quantifies disentanglement by averaging two metrics: \textit{Cause}, which measures how well the target attribute changes, and \textit{Isolation}, which ensures other attributes remain unaffected.




To summarize, we compare three SAE variants for each model: (1) the original pretrained SAE trained on the unpruned model, (2) the SAE retrained on the pruned model, and (3) the pruned version of the pretrained SAE. In all cases, we evaluate performance on \textit{WANDA}-pruned models to isolate the effects of SAE adaptation and pruning. Evaluation is conducted using SAEBench \citep{SAEBENCH}, which provides a standardized suite of metrics and tasks designed to assess the functional interpretability, reconstruction quality, and feature extraction capabilities of SAEs.

\section{Results}

\subsection{GPT-2 Small}
We conduct a two-stage experiment on GPT-2 Small. First, we apply WANDA pruning to the model and train an SAE on the resulting pruned model. In the second stage, we directly prune the pre-trained SAE trained on the original GPT-2 Small. Please refer to Section \ref{Exp setup: Model config} for a detailed description of our pruning setup. Finally, we compare the reconstruction loss of the trained SAE and its pruned counterpart on a held-out validation set.
\begin{figure}[htbp]
    \centering
    \begin{subfigure}[b]{0.32\textwidth}
        \includegraphics[width=\textwidth]{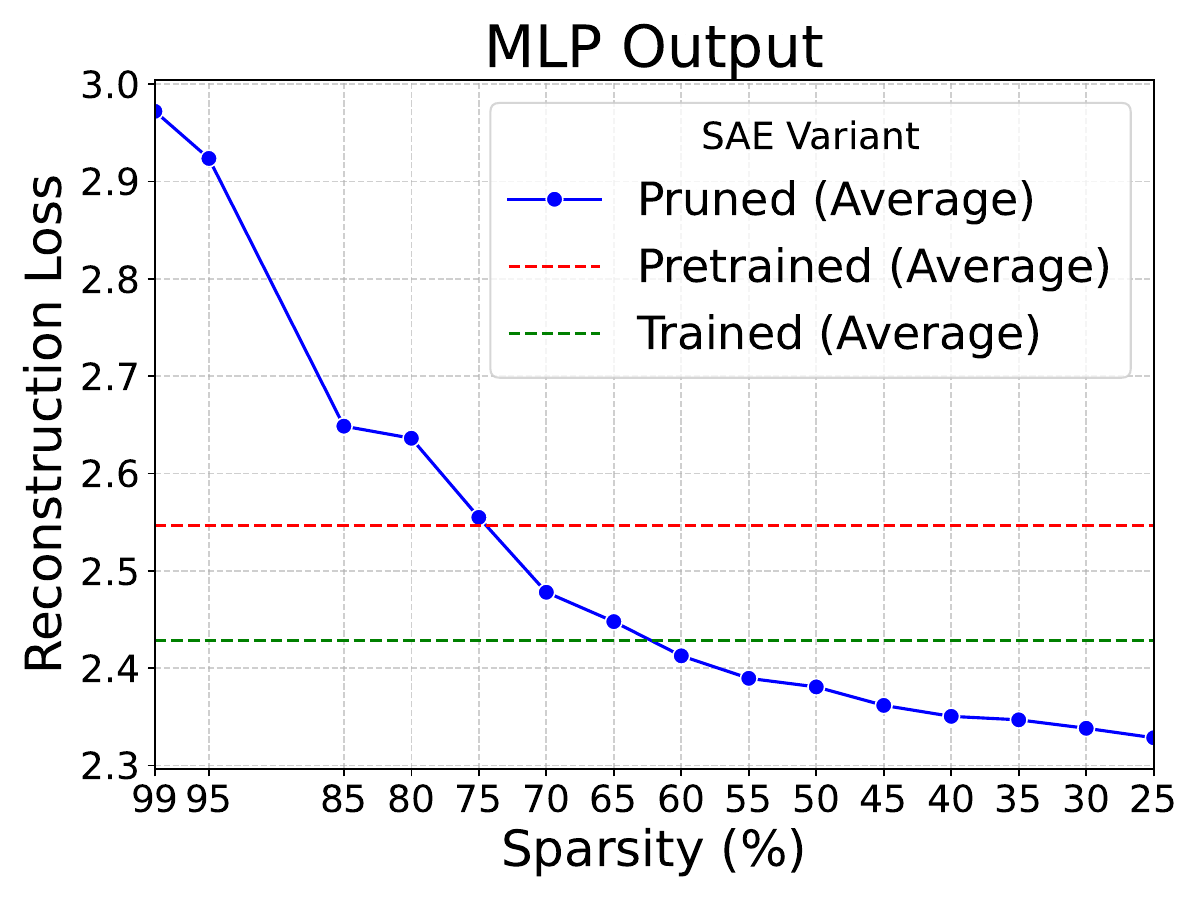}
        \label{fig:sub1}
    \end{subfigure}
    \hfill
    \begin{subfigure}[b]{0.32\textwidth}
        \includegraphics[width=\textwidth]{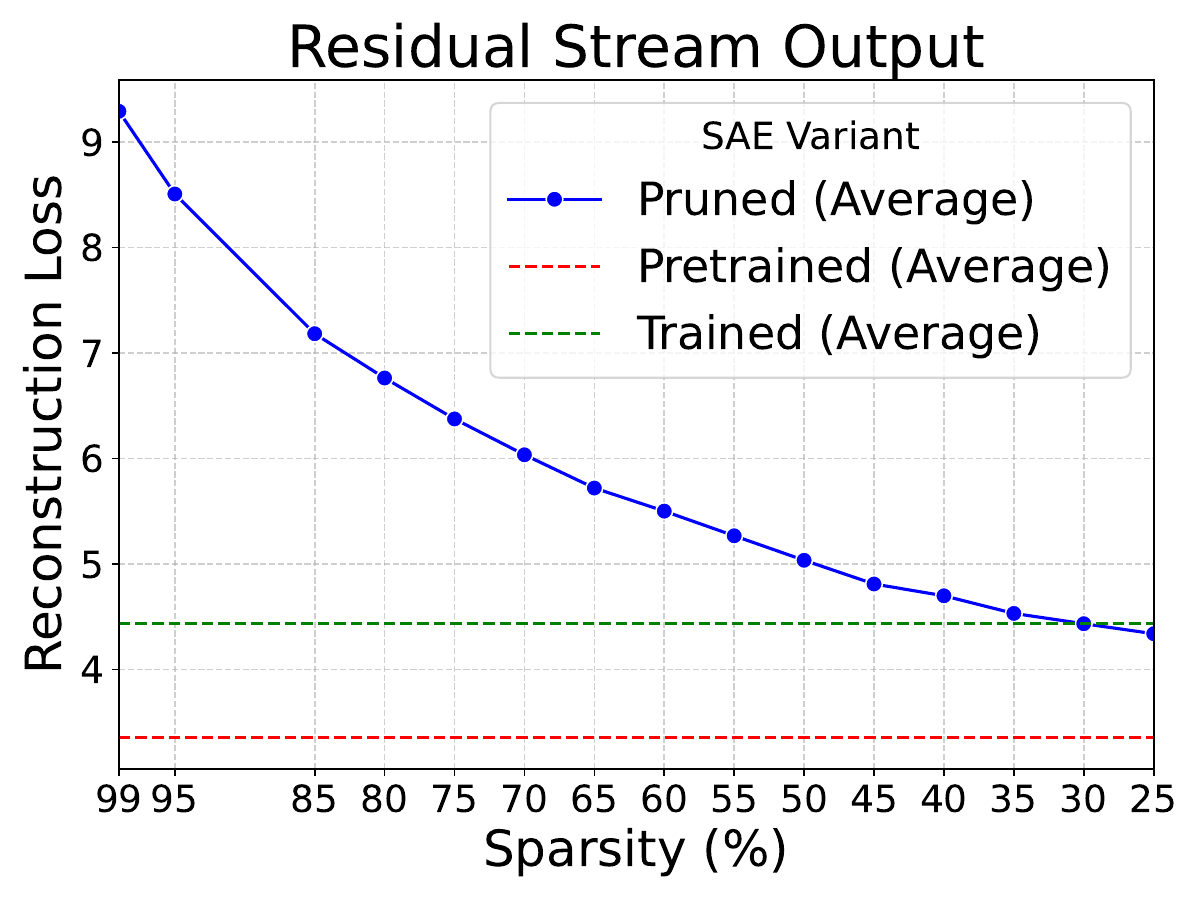}
        \label{fig:sub2}
    \end{subfigure}
    \hfill
    \begin{subfigure}[b]{0.32\textwidth}
        \includegraphics[width=\textwidth]{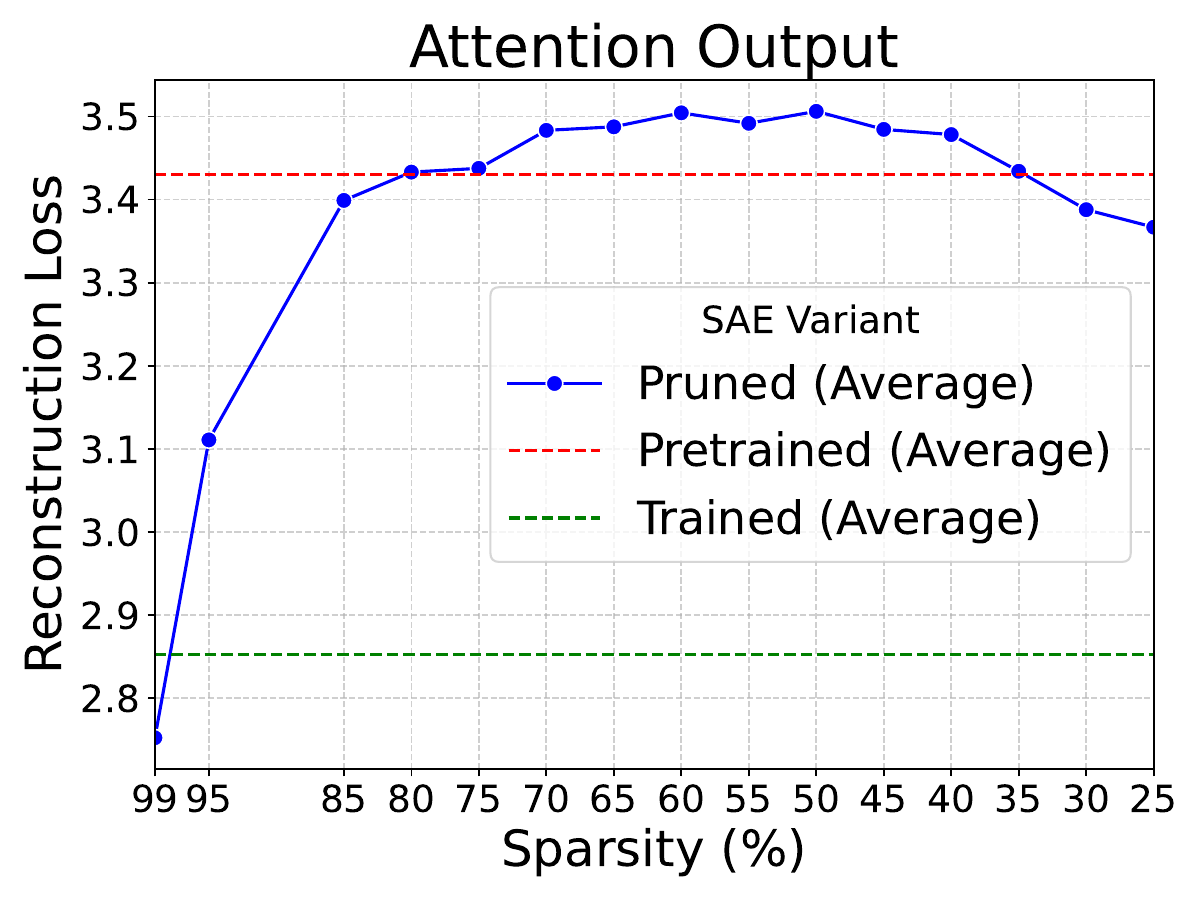}
        \label{fig:sub3}
    \end{subfigure}
    \caption{Average reconstruction loss of different SAE variants across all layers of GPT-2 Small at varying sparsity rates, ranging from 25\% to 99\%. Lower reconstruction loss indicates better preservation of information after WANDA pruning. Please refer to Appendix \ref{A1} for results on all layers of GPT-2 small.}
    \label{fig: recos loss gpt2-small}
\end{figure}

Figure \ref{fig: recos loss gpt2-small} illustrates SAE's reconstruction loss based on activation values of pruned GPT-2 as a function of sparsity rate. In this figure, we observe that across all output types, the pruned SAEs retain reconstruction performance comparable to their pretrained and fully trained counterparts. In particular, pruning up to 90\% sparsity leads to only a slight increase in reconstruction loss for the MLP and Residual Stream outputs, indicating that these components remain robust even after significant pruning. While the Attention output is somewhat more sensitive, performance remains stable up to around 85\% sparsity, suggesting that meaningful information is still preserved. We also want to emphasize that based on Figure \ref{fig: recos loss gpt2-small} and Appendix \ref{Section: Appendix eval metric cfg}, the pruned SAEs work better for MLP output and Attention output as compared to Residual Stream. 

These results demonstrate that sparse SAEs can effectively reconstruct activations even after aggressive pruning, supporting the potential for efficient interpretability in smaller models like GPT-2 Small. To test whether these trends generalize to larger language models with different internal structures and redundancy levels, we extend our analysis to \textbf{Gemma-2-2B}, a more modern and significantly larger architecture.

\subsection{Gemma-2-2B}
We evaluate four SAE variants on the Gemma-2-2B model. The \textbf{Pretrained} SAE is trained on the original, uncompressed model. The \textbf{Pruned25} and \textbf{Pruned50} variants refer to the same pretrained SAE pruned to 25\% and 50\% sparsity, respectively. The \textbf{Trained} SAE is trained from scratch on activations from the pruned model. These variants are assessed across a broad range of metrics from SAEBench \citep{SAEBENCH}, including absorption behavior, reconstruction fidelity, spurious correlation removal (SCR), targeted perturbation performance (TPP), and semantic disentanglement (RAVEL). Please refer to Section \ref{Exp setup: Eval metric config} for detailed descriptions of these evaluation metrics.

\begin{figure}[t]
\begin{subfigure}{0.5\textwidth}
\includegraphics[width=\linewidth]{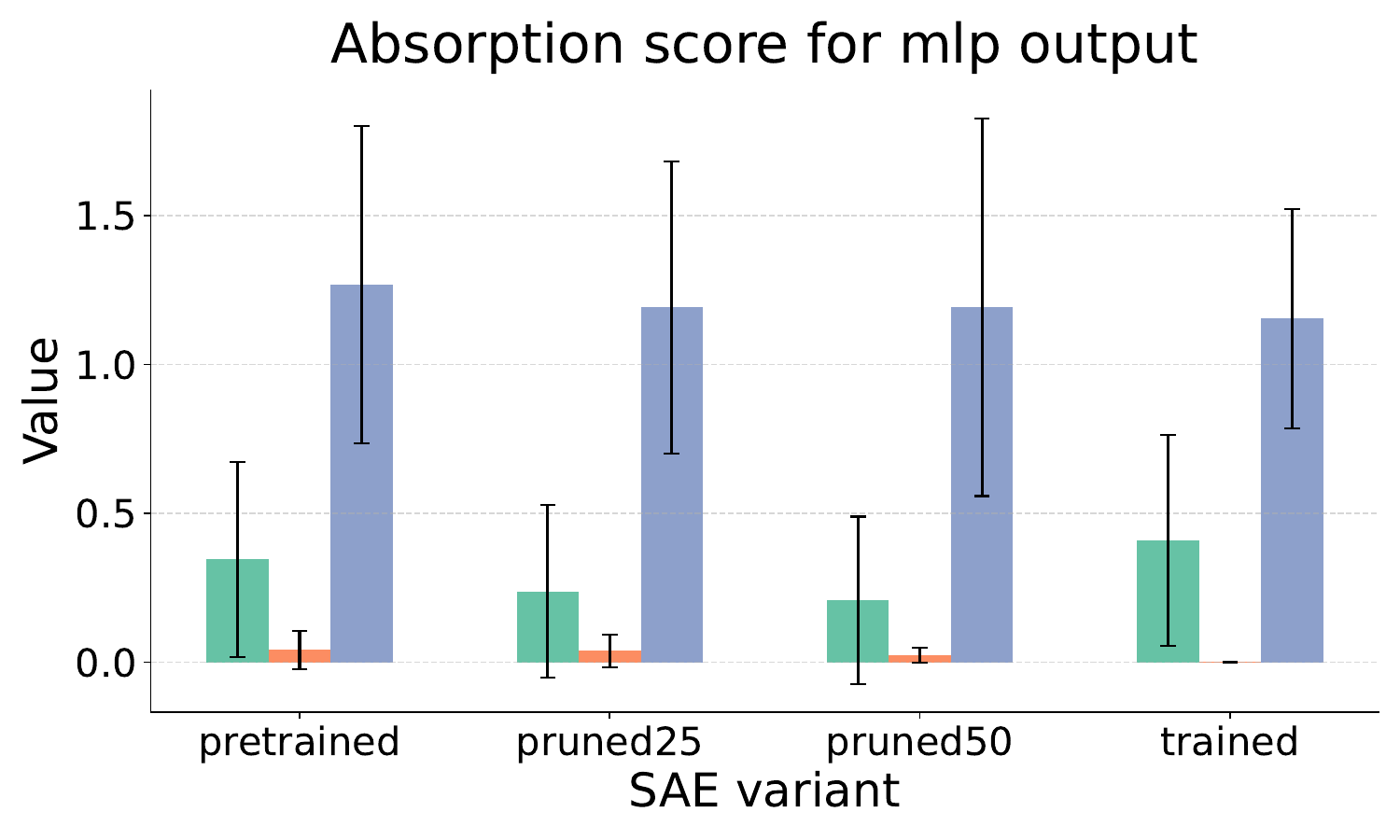} 
\label{fig:absorption_1}
\end{subfigure}
\begin{subfigure}{0.5\textwidth}
\includegraphics[width=\linewidth]{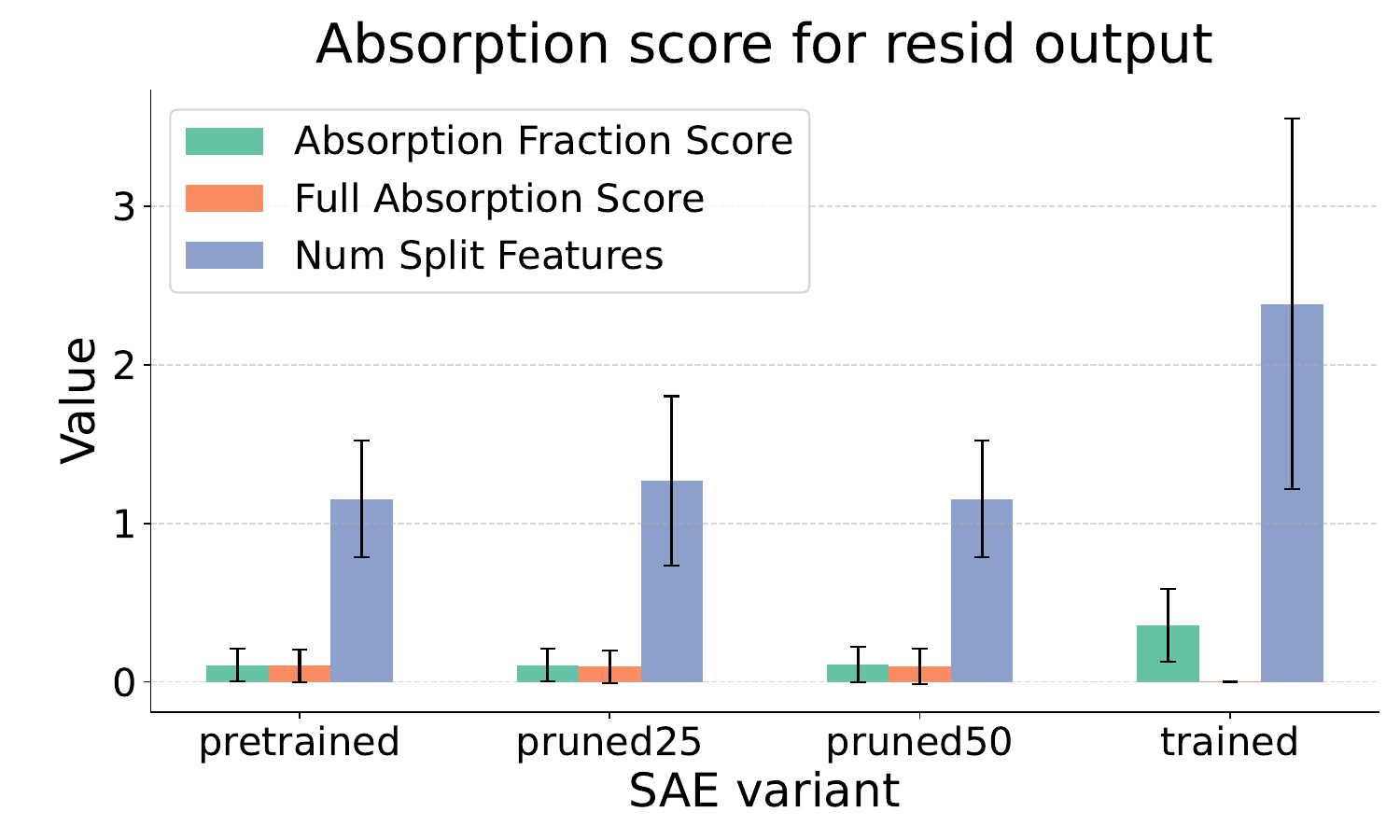}
\label{fig:absorption_2}
\end{subfigure}
\caption{Absorption scores for Gemma-2-2B SAE variants. Higher Absorption and Full Absorption scores indicate better interpretability, while fewer Split Features indicate more coherent and interpretable representations. Please refer Appendix \ref{A2} for more insights.}
\label{fig:Absorption score}
\end{figure}

Figure \ref{fig:Absorption score} reports absorption metrics at the MLP and residual layers across all model variants. Although \textit{Pruned25} and \textit{Pruned50} SAEs exhibit a minimal decline in absorption relative to the \textit{Pretrained} SAE, all of these variations of SAE achieve absorption scores comparable to those of the  \textit{Trained} SAE. These findings suggest that moderate pruning retains the SAE’s ability to recover abstract and structured features, reinforcing the hypothesis that pruned SAEs can retain key representational capabilities.

Table 1 summarizes the reconstruction performance of the four SAE variants, evaluated across attention, MLP, and residual outputs using KL divergence, cross entropy loss, cosine similarity, mean squared error, and explained variance. Across all output types, the both \textit{Pruned25} SAE and \textit{Pretrained} SAE have comparable performance to the \textit{Trained} SAE, with only modest reductions in reconstruction quality. More specifically, for \textit{Pruned25} SAE, the drop in cosine similarity and explained variance typically remains within a narrow 2 to 5\% range, indicating that the pruned SAE continues to capture the essential structure of the representations. As for the attention output, the reconstructed vectors remain similar, while for the MLP output, remain consistent. The residual component output, which reflects a combination of earlier computations, shows minimal degradation.

These findings indicate that pruning a pretrained SAE to 25\% sparsity retains most of its representational capacity. Indicating that training  an SAE from scratch for a pruned model is not necessary and a performant SAE can be attained via simply pruning the original SAE. Furthermore, this indicates that a small, structured subset of its neurons contributes to a substantial portion of the SAE’s reconstruction. This finding illustrates that benefits of direct pruning as a scalable and efficient approach to maintaining interpretability and reconstruction fidelity for SAE's.

\begin{table}[h]
\centering
\caption{
Core evaluation metrics across three output types—Attention, MLP, and Post-Residual—for the original model representations as well as four SAE variants: the \textbf{Pretrained} SAE, SAE pruned to 25\% sparsity (\textbf{Pruned25}), SAE pruned to 50\% sparsity (\textbf{Pruned50}), and an SAE trained directly on the WANDA-pruned model (\textbf{Trained}). The metrics include KL divergence (\textbf{kl div score}), cross-entropy loss (\textbf{ce loss score}), mean squared error (\textbf{mse}), and cosine similarity (\textbf{cossim}). All evaluations are conducted on \textbf{layer 12} of the \texttt{Gemma-2-2b} model. Further details regarding the metric configurations can be found in Appendix~\ref{A3}.}
\label{tab:core_outputs_combined}
{
\begin{tabular}{l|l|cccc}
\toprule
\textbf{Output} & \textbf{Metric} & \textbf{Pretrained} & \textbf{Pruned25} & \textbf{Pruned50} & \textbf{Trained} \\
\midrule
\multirow{8}{*}{Attention} 
& kl div score & 0.852 & 0.821 & 0.775 & 0.824 \\
& kl div with sae & 0.006 & 0.008 & 0.010 & 0.008 \\
& ce loss score & 0.862 & 1.000 & 0.667 & 1.000 \\
& ce loss with sae & 3.129 & 3.125 & 3.141 & 3.125 \\
& ce loss without sae & 3.124 & 3.125 & 3.125 & 3.125 \\
& explained variance & 0.802 & 0.793 & 0.727 & 0.859 \\
& mse & 0.035 & 0.037 & 0.049 & 0.025 \\
& cossim & 0.893 & 0.891 & 0.855 & 0.930 \\
\midrule
\multirow{8}{*}{MLP}
& kl div score & 0.777 & 0.714 & 0.604 & 0.502 \\
& kl div with sae & 0.043 & 0.015 & 0.021 & 0.026 \\
& ce loss score & 0.848 & 0.667 & 0.667 & 0.667 \\
& ce loss with sae & 5.085 & 3.141 & 3.141 & 3.141 \\
& ce loss without sae & 5.056 & 3.125 & 3.125 & 3.125 \\
& explained variance & 0.631 & 0.633 & 0.539 & 0.750 \\
& mse & 0.546 & 0.508 & 0.645 & 0.352 \\
& cossim & 0.787 & 0.789 & 0.730 & 0.867 \\
\midrule
\multirow{8}{*}{Post Residual}
& kl div score & 0.989 & 0.989 & 0.981 & 0.978 \\
& kl div with sae & 0.114 & 0.112 & 0.189 & 0.219 \\
& ce loss score & 0.988 & 0.990 & 0.982 & 0.982 \\
& ce loss with sae & 3.234 & 3.219 & 3.297 & 3.297 \\
& ce loss without sae & 3.125 & 3.125 & 3.125 & 3.125 \\
& explained variance & 0.867 & 0.867 & 0.828 & 0.883 \\
& mse & 1.680 & 1.703 & 2.156 & 1.492 \\
& cossim & 0.918 & 0.918 & 0.898 & 0.938 \\
\bottomrule
\end{tabular}}
\end{table}

\begin{figure}[h]
\begin{subfigure}{0.5\textwidth}
\includegraphics[width=\linewidth]{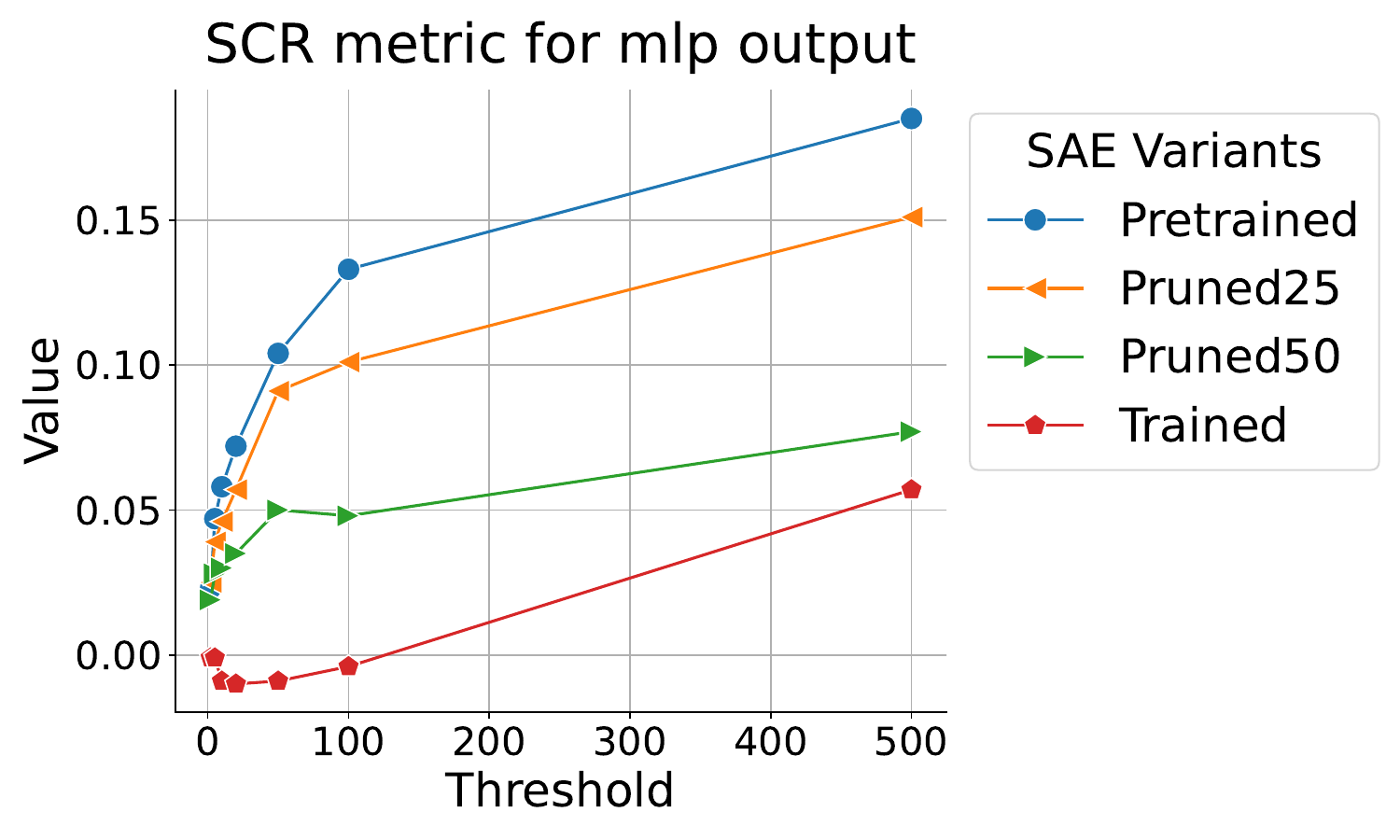} 
\label{fig: scr1}
\end{subfigure}
\begin{subfigure}{0.5\textwidth}
\includegraphics[width=\linewidth]{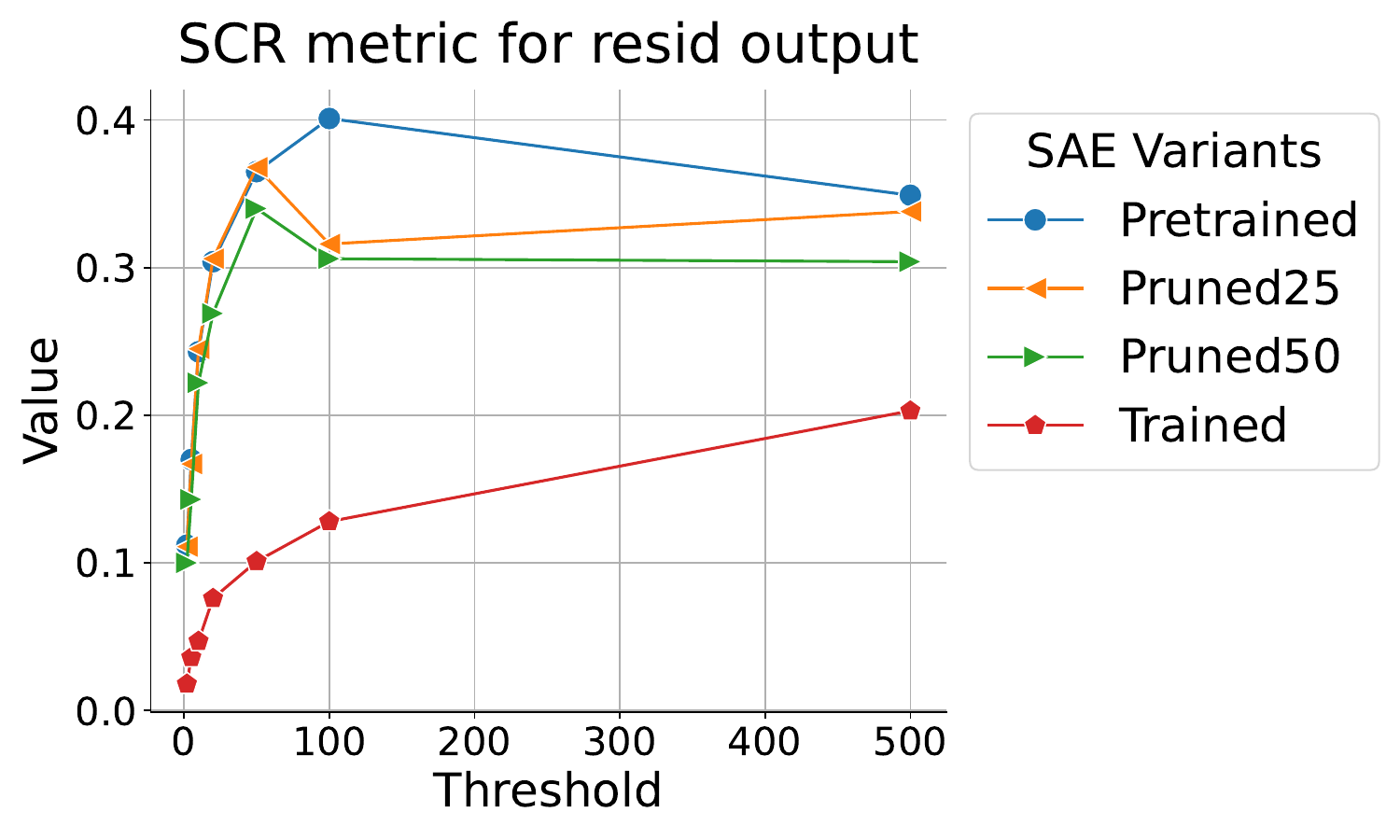}
\label{fig: scr2}
\end{subfigure}
\caption{SCR scores for Gemma-2-2B SAE variants. SCR measures the SAE's ability to eliminate features tied to spurious correlations. Higher scores reflect better removal of misleading latent patterns. Please refer Appendix \ref{A4} for a description of each metric.}
\label{fig: SCR metric}
\end{figure}

\begin{figure}[h]
\begin{subfigure}{0.5\textwidth}
\includegraphics[width=\linewidth]{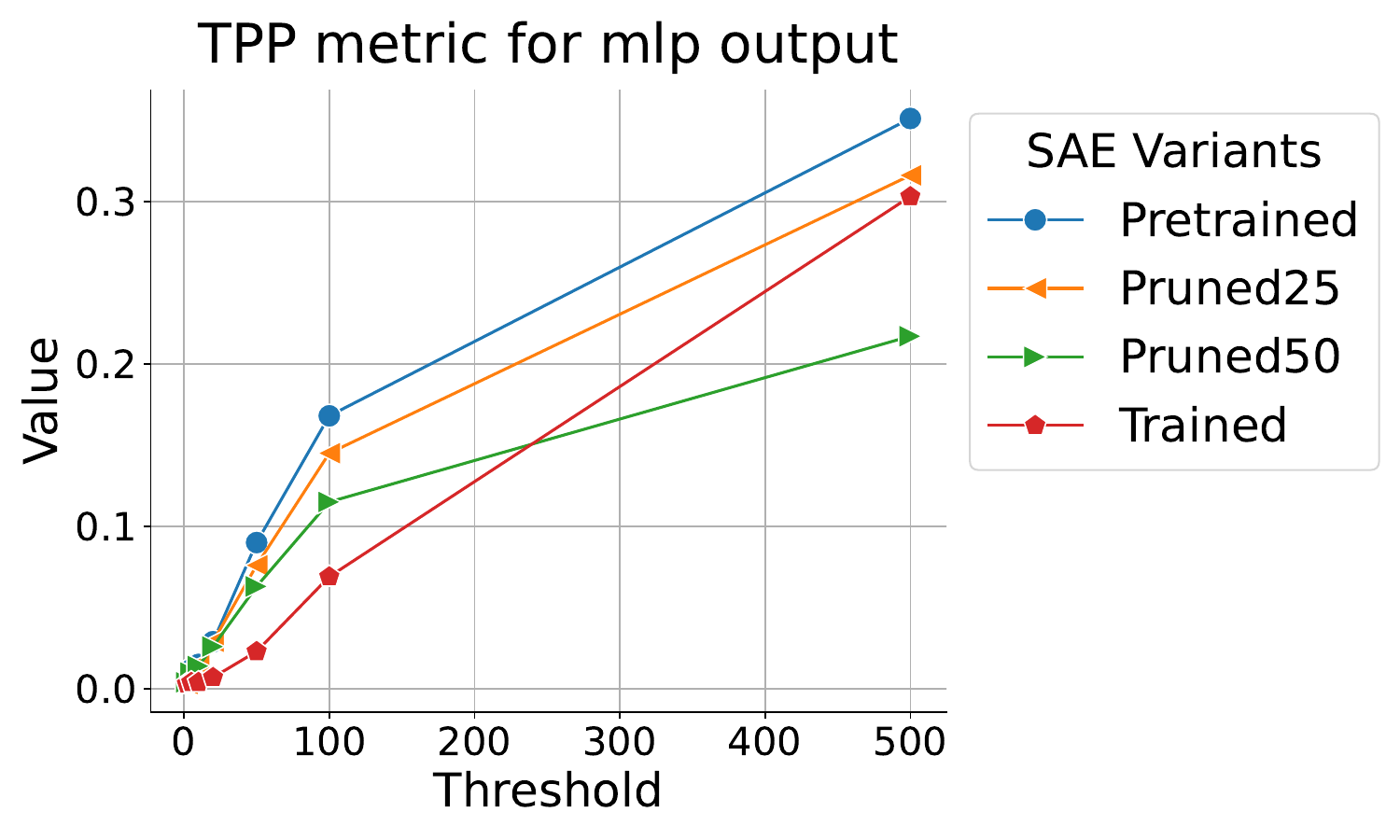} 
\label{fig: tpp1}
\end{subfigure}
\begin{subfigure}{0.5\textwidth}
\includegraphics[width=\linewidth]{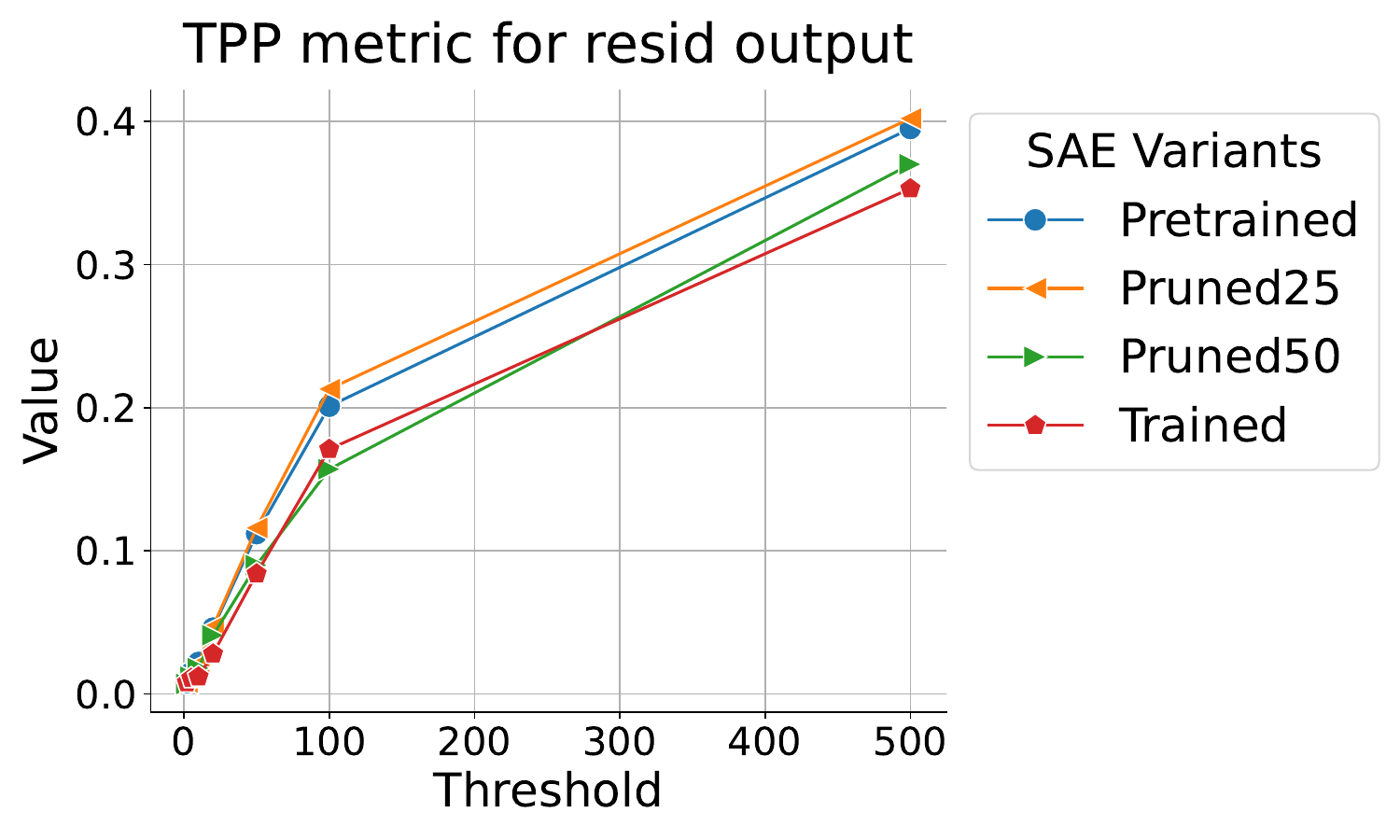}
\label{fig: tpp2}
\end{subfigure}
\caption{TPP scores for Gemma-2-2B SAE variants. TPP evaluates how well each SAE localizes concept-specific features. Higher TPP scores imply stronger disentanglement (ablating latents related to one class should mostly affect only that class). Please see Appendix \ref{A5} for more insights on the computation of the TPP score.}
\label{fig: TPP metric}
\end{figure}

Figures \ref{fig: SCR metric} and \ref{fig: TPP metric} evaluate the ability of SAEs to retain functional interpretability after pruning, focusing on two core aspects: Spurious Correlation Removal (SCR) and Targeted Probe Perturbation (TPP). Higher SCR and TPP implies that the SAE is more capable in removing correlations leading to bias.  We can observe that, across these metrics, both  \textit{Pruned25} SAE 
 and \textit{Pretrained} SAE  achieve highest SCR and TPP.
 This reinforces the idea that the reconstruction fidelity  of SAEs can be preserved without retraining, making Pretrained SAEs a practical and efficient option for downstream use.
 
More details on TTP and SCR can be found in  Tables \ref{tab_appendix:resid scr}, \ref{tab_appendix:mlp scr}, \ref{tab_appendix:mlp tpp} and \ref{tab_appendix:resid tpp} in  Appendix \ref{Section: Appendix eval metric cfg}. 

Finally, we evaluate semantic disentanglement using RAVEL, as summarized in Tables \ref{ravel tab: 1}, and \ref{ravel tab: 2}. Generally speaking, a higher RAVEL score implies that the intervention of SAE effectively changes the intended attribute, keeping the unrelated attributes unchanged (the higher value implies a better SAE). The \textit{Pretrained} and \textit{Pruned25}  SAEs consistently demonstrate stronger disentanglement than its Trained counterpart for the MLP output, whereas for Residual stream output the performances are comparable.  This improvement is especially apparent in challenging attribute pairs, where pruning appears to clarify latent representations. These results suggest that pruning can act as an implicit regularizer, removing noisy components while preserving the core semantic structure.

\begin{table}[h]
\centering
\caption{RAVEL disentanglement scores of Residual stream output for Gemma-2-2B SAE variants. Higher scores indicate better separation of semantic attributes.(See Appendix \ref{A6})}
\begin{tabular}{l|cccc}
\toprule
\textbf{Entity-Attribute pairs} & \textbf{Pretrained} & \textbf{Pruned25} & \textbf{Pruned50} & \textbf{Trained} \\
\midrule
(City, Country) & 0.5825 & 0.5629 & 0.6034 & 0.6418 \\
\midrule
(City, Continent) & 0.6007 & 0.6179 & 0.6139 & 0.6721 \\
\midrule
(City, Language) & 0.5665 & 0.5631 & 0.5566 & 0.5781 \\
\midrule
(Nobel prize winner, Birth Country) & 0.7922 & 0.8042 & 0.7440 & 0.6316 \\
\midrule
(Nobel prize winner, Field) & 0.8730 & 0.8724 & 0.8530 & 0.7969 \\
\midrule
(Nobel prize winner, Gender) & 0.7785 & 0.7973 & 0.7236 & 0.7148 \\
\bottomrule
\end{tabular}
\label{ravel tab: 1}
\end{table}

\begin{table}[h]
\centering
\caption{RAVEL disentanglement scores of MLP output for Gemma-2-2B SAE variants. Higher scores indicate better separation of semantic attributes. (See Appendix \ref{A7})}
\begin{tabular}{l|cccc}
\toprule
\textbf{Entity-Attribute pairs} & \textbf{Pretrained} & \textbf{Pruned25} & \textbf{Pruned50} & \textbf{Trained} \\
\midrule
(City, Country) & 0.5451 & 0.4688 & 0.3310 & 0.1948 \\
\midrule
(City, Continent) & 0.5922 & 0.5365 & 0.4204 & 0.2262 \\
\midrule
(City, Language) & 0.4746 & 0.4243 & 0.3421 & 0.2378 \\
\midrule
(Nobel prize winner, Birth Country) & 0.5331 & 0.4923 & 0.4397 & 0.3926 \\
\midrule
(Nobel prize winner, Field) & 0.7059 & 0.6798 & 0.6298 & 0.5385 \\
\midrule
(Nobel prize winner, Gender) & 0.8338 & 0.8252 & 0.7669 & 0.5431 \\
\bottomrule
\end{tabular}
\label{ravel tab: 2}
\end{table}

In summary, the \textit{Pruned25} SAE consistently matches or outperforms the retrained SAE across all evaluation metrics. These findings support our hypothesis and further underscore the effectiveness of pruning. Notably, pruning emerges as a practical and computationally efficient alternative to retraining, especially when the original SAE already encodes meaningful feature abstractions.
\section{Conclusion}
In this work, we explored the transferability of Sparse Autoencoders (SAEs) for interpreting compressed language models, focusing on models pruned using the WANDA method. Our experiments across two architectures—GPT-2 Small and Gemma-2-2B—demonstrate that SAEs pretrained on uncompressed models retain high interpretability and reconstruction fidelity even when directly applied to pruned models. Moreover, simply pruning the pretrained SAE itself yields performance that closely matches that of SAEs retrained from scratch on pruned models. This observation holds across a wide range of metrics, including reconstruction loss, spurious correlation removal, semantic disentanglement, and feature absorption. These findings suggest that retraining SAEs for each compressed variant of a model may be unnecessary, offering a significant reduction in computational cost while preserving interpretability. Additionally, the success of pruned SAEs in maintaining key representational and behavioral features positions them as a practical tool for the post-hoc analysis of compressed models. More broadly, our results reinforce the potential for efficient and scalable interpretability frameworks that align with the growing need for resource-conscious model deployment. Future work may explore the generalization of these findings to quantized and low-rank compressed models and further investigate the role of pruning as an implicit regularizer in interpretability systems.

\section{Acknowledgment}
This work is supported by the U.S. National Science Foundation under award
IIS-2301599 and CMMI-2301601, and by grants from the Ohio State University’s Translational Data
Analytics Institute and College of Engineering Strategic Research Initiative.

\bibliography{colm2025_conference}

\begin{thebibliography}{47}
\providecommand{\natexlab}[1]{#1}
\providecommand{\url}[1]{\texttt{#1}}
\expandafter\ifx\csname urlstyle\endcsname\relax
  \providecommand{\doi}[1]{doi: #1}\else
  \providecommand{\doi}{doi: \begingroup \urlstyle{rm}\Url}\fi

\bibitem[Arditi et~al.(2024)Arditi, Obeso, Syed, Paleka, Panickssery, Gurnee, and Nanda]{arditi2024refusal}
Andy Arditi, Oscar Obeso, Aaquib Syed, Daniel Paleka, Nina Panickssery, Wes Gurnee, and Neel Nanda.
\newblock Refusal in language models is mediated by a single direction.
\newblock \emph{arXiv preprint arXiv:2406.11717}, 2024.

\bibitem[B. et~al.(2024a)B., P., and Nanda]{batchtopk}
Bussmann B., Leask P., and N~Nanda.
\newblock Batchtopk: A simple improvement for topk-saes, 2024a.
\newblock URL \url{https://www.alignmentforum.org/posts/rKM9b6B2LqwSB5ToN/learning-multi-level-features-with-matryoshka-saes}.

\bibitem[Bai et~al.(2022)Bai, Hou, Shang, Jiang, King, and Lyu]{bai2022towards}
Haoli Bai, Lu~Hou, Lifeng Shang, Xin Jiang, Irwin King, and Michael~R Lyu.
\newblock Towards efficient post-training quantization of pre-trained language models.
\newblock \emph{Advances in neural information processing systems}, 35:\penalty0 1405--1418, 2022.

\bibitem[Chanin et~al.(2024)Chanin, Wilken-Smith, Dulka, Bhatnagar, and Bloom]{ABSORPTION}
David Chanin, James Wilken-Smith, Tomáš Dulka, Hardik Bhatnagar, and Joseph Bloom.
\newblock A is for absorption: Studying feature splitting and absorption in sparse autoencoders, 2024.
\newblock URL \url{https://arxiv.org/abs/2409.14507}.

\bibitem[Chhabra et~al.(2025)Chhabra, Zhu, and Khalili]{chhabra2025neuroplasticity}
Vishnu~Kabir Chhabra, Ding Zhu, and Mohammad~Mahdi Khalili.
\newblock Neuroplasticity and corruption in model mechanisms: A case study of indirect object identification.
\newblock \emph{arXiv preprint arXiv:2503.01896}, 2025.

\bibitem[Conmy et~al.(2023)Conmy, Mavor-Parker, Lynch, Heimersheim, and Garriga-Alonso]{conmy2023towards}
Arthur Conmy, Augustine Mavor-Parker, Aengus Lynch, Stefan Heimersheim, and Adri{\`a} Garriga-Alonso.
\newblock Towards automated circuit discovery for mechanistic interpretability.
\newblock \emph{Advances in Neural Information Processing Systems}, 36:\penalty0 16318--16352, 2023.

\bibitem[Cunningham et~al.(2023)Cunningham, Ewart, Riggs, Huben, and Sharkey]{cunningham2023sparse}
Hoagy Cunningham, Aidan Ewart, Logan Riggs, Robert Huben, and Lee Sharkey.
\newblock Sparse autoencoders find highly interpretable features in language models.
\newblock \emph{arXiv preprint arXiv:2309.08600}, 2023.

\bibitem[Frantar \& Alistarh(2023{\natexlab{a}})Frantar and Alistarh]{frantar2023sparsegpt}
Elias Frantar and Dan Alistarh.
\newblock Sparsegpt: Massive language models can be accurately pruned in one-shot.
\newblock In \emph{International Conference on Machine Learning}, pp.\  10323--10337. PMLR, 2023{\natexlab{a}}.

\bibitem[Frantar \& Alistarh(2023{\natexlab{b}})Frantar and Alistarh]{sparsegpt}
Elias Frantar and Dan Alistarh.
\newblock Sparsegpt: Massive language models can be accurately pruned in one-shot, 2023{\natexlab{b}}.
\newblock URL \url{https://arxiv.org/abs/2301.00774}.

\bibitem[Gao et~al.(2020)Gao, Biderman, Black, Golding, Hoppe, Foster, Phang, He, Thite, Nabeshima, Presser, and Leahy]{gao2020pile800gbdatasetdiverse}
Leo Gao, Stella Biderman, Sid Black, Laurence Golding, Travis Hoppe, Charles Foster, Jason Phang, Horace He, Anish Thite, Noa Nabeshima, Shawn Presser, and Connor Leahy.
\newblock The pile: An 800gb dataset of diverse text for language modeling, 2020.
\newblock URL \url{https://arxiv.org/abs/2101.00027}.

\bibitem[Gao et~al.(2024{\natexlab{a}})Gao, la~Tour, Tillman, Goh, Troll, Radford, Sutskever, Leike, and Wu]{gao2024scaling}
Leo Gao, Tom~Dupr{\'e} la~Tour, Henk Tillman, Gabriel Goh, Rajan Troll, Alec Radford, Ilya Sutskever, Jan Leike, and Jeffrey Wu.
\newblock Scaling and evaluating sparse autoencoders.
\newblock \emph{arXiv preprint arXiv:2406.04093}, 2024{\natexlab{a}}.

\bibitem[Gao et~al.(2024{\natexlab{b}})Gao, la~Tour, Tillman, Goh, Troll, Radford, Sutskever, Leike, and Wu]{topk}
Leo Gao, Tom~Dupré la~Tour, Henk Tillman, Gabriel Goh, Rajan Troll, Alec Radford, Ilya Sutskever, Jan Leike, and Jeffrey Wu.
\newblock Scaling and evaluating sparse autoencoders, 2024{\natexlab{b}}.
\newblock URL \url{https://arxiv.org/abs/2406.04093}.

\bibitem[Gokaslan et~al.(2019)Gokaslan, Cohen, Pavlick, and Tellex]{Gokaslan2019OpenWeb}
Aaron Gokaslan, Vanya Cohen, Ellie Pavlick, and Stefanie Tellex.
\newblock Openwebtext corpus.
\newblock \url{http://Skylion007.github.io/OpenWebTextCorpus}, 2019.

\bibitem[Guo et~al.(2025)Guo, Yang, Zhang, Song, Zhang, Xu, Zhu, Ma, Wang, Bi, et~al.]{guo2025deepseek}
Daya Guo, Dejian Yang, Haowei Zhang, Junxiao Song, Ruoyu Zhang, Runxin Xu, Qihao Zhu, Shirong Ma, Peiyi Wang, Xiao Bi, et~al.
\newblock Deepseek-r1: Incentivizing reasoning capability in llms via reinforcement learning.
\newblock \emph{arXiv preprint arXiv:2501.12948}, 2025.

\bibitem[Gurnee et~al.(2023)Gurnee, Nanda, Pauly, Harvey, Troitskii, and Bertsimas]{gurnee2023findingneuronshaystackcase}
Wes Gurnee, Neel Nanda, Matthew Pauly, Katherine Harvey, Dmitrii Troitskii, and Dimitris Bertsimas.
\newblock Finding neurons in a haystack: Case studies with sparse probing, 2023.
\newblock URL \url{https://arxiv.org/abs/2305.01610}.

\bibitem[Han et~al.(2015)Han, Pool, Tran, and Dally]{magnitude}
Song Han, Jeff Pool, John Tran, and William~J. Dally.
\newblock Learning both weights and connections for efficient neural networks, 2015.
\newblock URL \url{https://arxiv.org/abs/1506.02626}.

\bibitem[Hanna et~al.(2023)Hanna, Liu, and Variengien]{hanna2023does}
Michael Hanna, Ollie Liu, and Alexandre Variengien.
\newblock How does gpt-2 compute greater-than?: Interpreting mathematical abilities in a pre-trained language model.
\newblock \emph{Advances in Neural Information Processing Systems}, 36:\penalty0 76033--76060, 2023.

\bibitem[Hong et~al.(2024)Hong, Duan, Zhang, Li, Xie, Lieberman, Diffenderfer, Bartoldson, Jaiswal, Xu, et~al.]{hong2024decoding}
Junyuan Hong, Jinhao Duan, Chenhui Zhang, Zhangheng Li, Chulin Xie, Kelsey Lieberman, James Diffenderfer, Brian Bartoldson, Ajay Jaiswal, Kaidi Xu, et~al.
\newblock Decoding compressed trust: Scrutinizing the trustworthiness of efficient llms under compression.
\newblock \emph{arXiv preprint arXiv:2403.15447}, 2024.

\bibitem[Hsu et~al.(2022)Hsu, Hua, Chang, Lou, Shen, and Jin]{hsu2022languagemodelcompressionweighted}
Yen-Chang Hsu, Ting Hua, Sungen Chang, Qian Lou, Yilin Shen, and Hongxia Jin.
\newblock Language model compression with weighted low-rank factorization, 2022.
\newblock URL \url{https://arxiv.org/abs/2207.00112}.

\bibitem[Joseph~Bloom \& Chanin(2024)Joseph~Bloom and Chanin]{bloom2024saetrainingcodebase}
Curt~Tigges Joseph~Bloom and David Chanin.
\newblock Saelens.
\newblock \url{https://github.com/jbloomAus/SAELens}, 2024.

\bibitem[Karvonen et~al.(2025{\natexlab{a}})Karvonen, Rager, Lin, Tigges, Bloom, Chanin, Lau, Farrell, McDougall, Ayonrinde, Wearden, Conmy, Marks, and Nanda]{SAEBENCH}
Adam Karvonen, Can Rager, Johnny Lin, Curt Tigges, Joseph Bloom, David Chanin, Yeu-Tong Lau, Eoin Farrell, Callum McDougall, Kola Ayonrinde, Matthew Wearden, Arthur Conmy, Samuel Marks, and Neel Nanda.
\newblock Saebench: A comprehensive benchmark for sparse autoencoders in language model interpretability, 2025{\natexlab{a}}.
\newblock URL \url{https://arxiv.org/abs/2503.09532}.

\bibitem[Karvonen et~al.(2025{\natexlab{b}})Karvonen, Rager, Lin, Tigges, Bloom, Chanin, Lau, Farrell, McDougall, Ayonrinde, Wearden, Conmy, Marks, and Nanda]{SAEBENCH_blog}
Adam Karvonen, Can Rager, Johnny Lin, Curt Tigges, Joseph Bloom, David Chanin, Yeu-Tong Lau, Eoin Farrell, Callum McDougall, Kola Ayonrinde, Matthew Wearden, Arthur Conmy, Samuel Marks, and Neel Nanda.
\newblock Saebench: A comprehensive benchmark for sparse autoencoders in language model interpretability, 2025{\natexlab{b}}.
\newblock URL \url{https://www.neuronpedia.org/sae-bench/info#introduction}.

\bibitem[Lan et~al.(2024)Lan, Torr, Meek, Khakzar, Krueger, and Barez]{lan2024sparse}
Michael Lan, Philip Torr, Austin Meek, Ashkan Khakzar, David Krueger, and Fazl Barez.
\newblock Sparse autoencoders reveal universal feature spaces across large language models.
\newblock \emph{arXiv preprint arXiv:2410.06981}, 2024.

\bibitem[Lee~Sharkey(2022)]{leeshark}
Beren~Millidge Lee~Sharkey, Dan~Braun.
\newblock Taking features out of superposition with sparse autoencoders, 2022.
\newblock URL \url{https://www.alignmentforum.org/posts/z6QQJbtpkEAX3Aojj/interim-research-report-taking-features-out-of-superposition}.

\bibitem[Lieberum et~al.(2024)Lieberum, Rajamanoharan, Conmy, Smith, Sonnerat, Varma, Kramár, Dragan, Shah, and Nanda]{lieberum2024gemmascope}
Tom Lieberum, Senthooran Rajamanoharan, Arthur Conmy, Lewis Smith, Nicolas Sonnerat, Vikrant Varma, János Kramár, Anca Dragan, Rohin Shah, and Neel Nanda.
\newblock Gemma scope: Open sparse autoencoders everywhere all at once on gemma 2, 2024.
\newblock URL \url{https://arxiv.org/abs/2408.05147}.

\bibitem[Lin et~al.(2024)Lin, Tang, Tang, Yang, Chen, Wang, Xiao, Dang, Gan, and Han]{lin2024awq}
Ji~Lin, Jiaming Tang, Haotian Tang, Shang Yang, Wei-Ming Chen, Wei-Chen Wang, Guangxuan Xiao, Xingyu Dang, Chuang Gan, and Song Han.
\newblock Awq: Activation-aware weight quantization for on-device llm compression and acceleration.
\newblock \emph{Proceedings of Machine Learning and Systems}, 6:\penalty0 87--100, 2024.

\bibitem[Ma et~al.(2023)Ma, Fang, and Wang]{ma2023llm}
Xinyin Ma, Gongfan Fang, and Xinchao Wang.
\newblock Llm-pruner: On the structural pruning of large language models.
\newblock \emph{Advances in neural information processing systems}, 36:\penalty0 21702--21720, 2023.

\bibitem[Makelov et~al.(2024)Makelov, Lange, and Nanda]{makelov2024principledevaluationssparseautoencoders}
Aleksandar Makelov, George Lange, and Neel Nanda.
\newblock Towards principled evaluations of sparse autoencoders for interpretability and control, 2024.
\newblock URL \url{https://arxiv.org/abs/2405.08366}.

\bibitem[Marks et~al.(2024)Marks, Rager, Michaud, Belinkov, Bau, and Mueller]{SHIFT}
Samuel Marks, Can Rager, Eric~J. Michaud, Yonatan Belinkov, David Bau, and Aaron Mueller.
\newblock Sparse feature circuits: Discovering and editing interpretable causal graphs in language models, 2024.
\newblock URL \url{https://arxiv.org/abs/2403.19647}.

\bibitem[Nanda \& Bloom(2022)Nanda and Bloom]{nanda2022transformerlens}
Neel Nanda and Joseph Bloom.
\newblock Transformerlens.
\newblock \url{https://github.com/TransformerLensOrg/TransformerLens}, 2022.

\bibitem[Paulo \& Belrose(2025)Paulo and Belrose]{paulo2025sparseautoencoderstraineddata}
Gonçalo Paulo and Nora Belrose.
\newblock Sparse autoencoders trained on the same data learn different features, 2025.
\newblock URL \url{https://arxiv.org/abs/2501.16615}.

\bibitem[Prakash et~al.(2024)Prakash, Shaham, Haklay, Belinkov, and Bau]{prakash2024fine}
Nikhil Prakash, Tamar~Rott Shaham, Tal Haklay, Yonatan Belinkov, and David Bau.
\newblock Fine-tuning enhances existing mechanisms: A case study on entity tracking.
\newblock \emph{arXiv preprint arXiv:2402.14811}, 2024.

\bibitem[Radford et~al.(2019)Radford, Wu, Child, Luan, Amodei, and Sutskever]{gpt2}
Alec Radford, Jeff Wu, Rewon Child, David Luan, Dario Amodei, and Ilya Sutskever.
\newblock Language models are unsupervised multitask learners.
\newblock 2019.

\bibitem[Rajamanoharan et~al.(2024)Rajamanoharan, Lieberum, Sonnerat, Conmy, Varma, Kramár, and Nanda]{jumprelu}
Senthooran Rajamanoharan, Tom Lieberum, Nicolas Sonnerat, Arthur Conmy, Vikrant Varma, János Kramár, and Neel Nanda.
\newblock Jumping ahead: Improving reconstruction fidelity with jumprelu sparse autoencoders, 2024.
\newblock URL \url{https://arxiv.org/abs/2407.14435}.

\bibitem[Saha et~al.(2024)Saha, Sagan, Srivastava, Goldsmith, and Pilanci]{saha2024compressinglargelanguagemodels}
Rajarshi Saha, Naomi Sagan, Varun Srivastava, Andrea~J. Goldsmith, and Mert Pilanci.
\newblock Compressing large language models using low rank and low precision decomposition, 2024.
\newblock URL \url{https://arxiv.org/abs/2405.18886}.

\bibitem[Shao et~al.(2023)Shao, Chen, Zhang, Xu, Zhao, Li, Zhang, Gao, Qiao, and Luo]{shao2023omniquant}
Wenqi Shao, Mengzhao Chen, Zhaoyang Zhang, Peng Xu, Lirui Zhao, Zhiqian Li, Kaipeng Zhang, Peng Gao, Yu~Qiao, and Ping Luo.
\newblock Omniquant: Omnidirectionally calibrated quantization for large language models.
\newblock \emph{arXiv preprint arXiv:2308.13137}, 2023.

\bibitem[Sun et~al.(2023)Sun, Liu, Bair, and Kolter]{wanda}
Mingjie Sun, Zhuang Liu, Anna Bair, and J~Zico Kolter.
\newblock A simple and effective pruning approach for large language models.
\newblock \emph{arXiv preprint arXiv:2306.11695}, 2023.

\bibitem[T. et~al.(2023)T., A., J., B., Jermyn~A., N., C., C., Askell~A., Y., S., N., T., N., Z., A., K., B., E., T., Carter~S., and C.]{bricken2023interpreting}
Bricken T., Templeton A., Batson J., Chen B., Conerly~T. Jermyn~A., Turner N., Anil C., Denison C., Lasenby~R. Askell~A., Wu~Y., Kravec S., Schiefer N., Maxwell T., Joseph N., Hatfield-Dodds Z., Tamkin A., Nguyen K., McLean B., Burke~J. E., Hume T., Henighan~T. Carter~S., and Olah C.
\newblock Towards monosemanticity: Decomposing language models with dictionary learning, 2023.
\newblock URL \url{https://transformer-circuits.pub/2023/monosemantic-features}.

\bibitem[Taggart(2024)]{prolu}
G.~M.~Prolu Taggart.
\newblock A nonlinearity for sparse autoencoders, 2024.
\newblock URL \url{https://www.alignmentforum.org/posts/HEpufTdakGTTKgoYF/prolu-a-nonlinearity-for-sparse-autoencoders}.

\bibitem[Team et~al.(2024)Team, Riviere, Pathak, Sessa, Hardin, Bhupatiraju, Hussenot, Mesnard, Shahriari, Ram{\'e}, et~al.]{team2024gemma}
Gemma Team, Morgane Riviere, Shreya Pathak, Pier~Giuseppe Sessa, Cassidy Hardin, Surya Bhupatiraju, L{\'e}onard Hussenot, Thomas Mesnard, Bobak Shahriari, Alexandre Ram{\'e}, et~al.
\newblock Gemma 2: Improving open language models at a practical size.
\newblock \emph{arXiv preprint arXiv:2408.00118}, 2024.

\bibitem[Touvron et~al.(2023)Touvron, Martin, Stone, Albert, Almahairi, Babaei, Bashlykov, Batra, Bhargava, Bhosale, et~al.]{touvron2023llama}
Hugo Touvron, Louis Martin, Kevin Stone, Peter Albert, Amjad Almahairi, Yasmine Babaei, Nikolay Bashlykov, Soumya Batra, Prajjwal Bhargava, Shruti Bhosale, et~al.
\newblock Llama 2: Open foundation and fine-tuned chat models.
\newblock \emph{arXiv preprint arXiv:2307.09288}, 2023.

\bibitem[Vaswani et~al.(2017)Vaswani, Shazeer, Parmar, Uszkoreit, Jones, Gomez, Kaiser, and Polosukhin]{Vaswani+2017}
Ashish Vaswani, Noam Shazeer, Niki Parmar, Jakob Uszkoreit, Llion Jones, Aidan~N Gomez, \L~ukasz Kaiser, and Illia Polosukhin.
\newblock Attention is all you need.
\newblock In \emph{Advances in Neural Information Processing Systems}, volume~30. Curran Associates, Inc., 2017.
\newblock URL \url{https://proceedings.neurips.cc/paper_files/paper/2017/file/3f5ee243547dee91fbd053c1c4a845aa-Paper.pdf}.

\bibitem[Wang et~al.(2022)Wang, Variengien, Conmy, Shlegeris, and Steinhardt]{wang2022interpretability}
Kevin Wang, Alexandre Variengien, Arthur Conmy, Buck Shlegeris, and Jacob Steinhardt.
\newblock Interpretability in the wild: a circuit for indirect object identification in gpt-2 small.
\newblock \emph{arXiv preprint arXiv:2211.00593}, 2022.

\bibitem[Wright \& Ma(2022)Wright and Ma]{wright}
John Wright and Yi~Ma.
\newblock \emph{High-Dimensional Data Analysis with Low-Dimensional Models: Principles, Computation, and Applications}.
\newblock Cambridge University Press, 2022.

\bibitem[Xiao et~al.(2023)Xiao, Lin, Seznec, Wu, Demouth, and Han]{xiao2023smoothquant}
Guangxuan Xiao, Ji~Lin, Mickael Seznec, Hao Wu, Julien Demouth, and Song Han.
\newblock Smoothquant: Accurate and efficient post-training quantization for large language models.
\newblock In \emph{International Conference on Machine Learning}, pp.\  38087--38099. PMLR, 2023.

\bibitem[Xu et~al.(2024)Xu, Gupta, Li, Bentham, and Srikumar]{xu2024perplexitymultidimensionalsafetyevaluation}
Zhichao Xu, Ashim Gupta, Tao Li, Oliver Bentham, and Vivek Srikumar.
\newblock Beyond perplexity: Multi-dimensional safety evaluation of llm compression, 2024.
\newblock URL \url{https://arxiv.org/abs/2407.04965}.

\bibitem[Zhou et~al.(2024)Zhou, Ning, Hong, Fu, Xu, Li, Lou, Wang, Yuan, Li, et~al.]{zhou2024survey}
Zixuan Zhou, Xuefei Ning, Ke~Hong, Tianyu Fu, Jiaming Xu, Shiyao Li, Yuming Lou, Luning Wang, Zhihang Yuan, Xiuhong Li, et~al.
\newblock A survey on efficient inference for large language models.
\newblock \emph{arXiv preprint arXiv:2404.14294}, 2024.

\end{thebibliography}
\bibliographystyle{colm2025_conference}

\appendix
\section{Evaluation metric configuration} \label{Section: Appendix eval metric cfg}

\subsection{Exploratory analysis on GPT-2 small} \label{A1}
\begin{figure*}
    \centering
    \includegraphics[width=\linewidth]{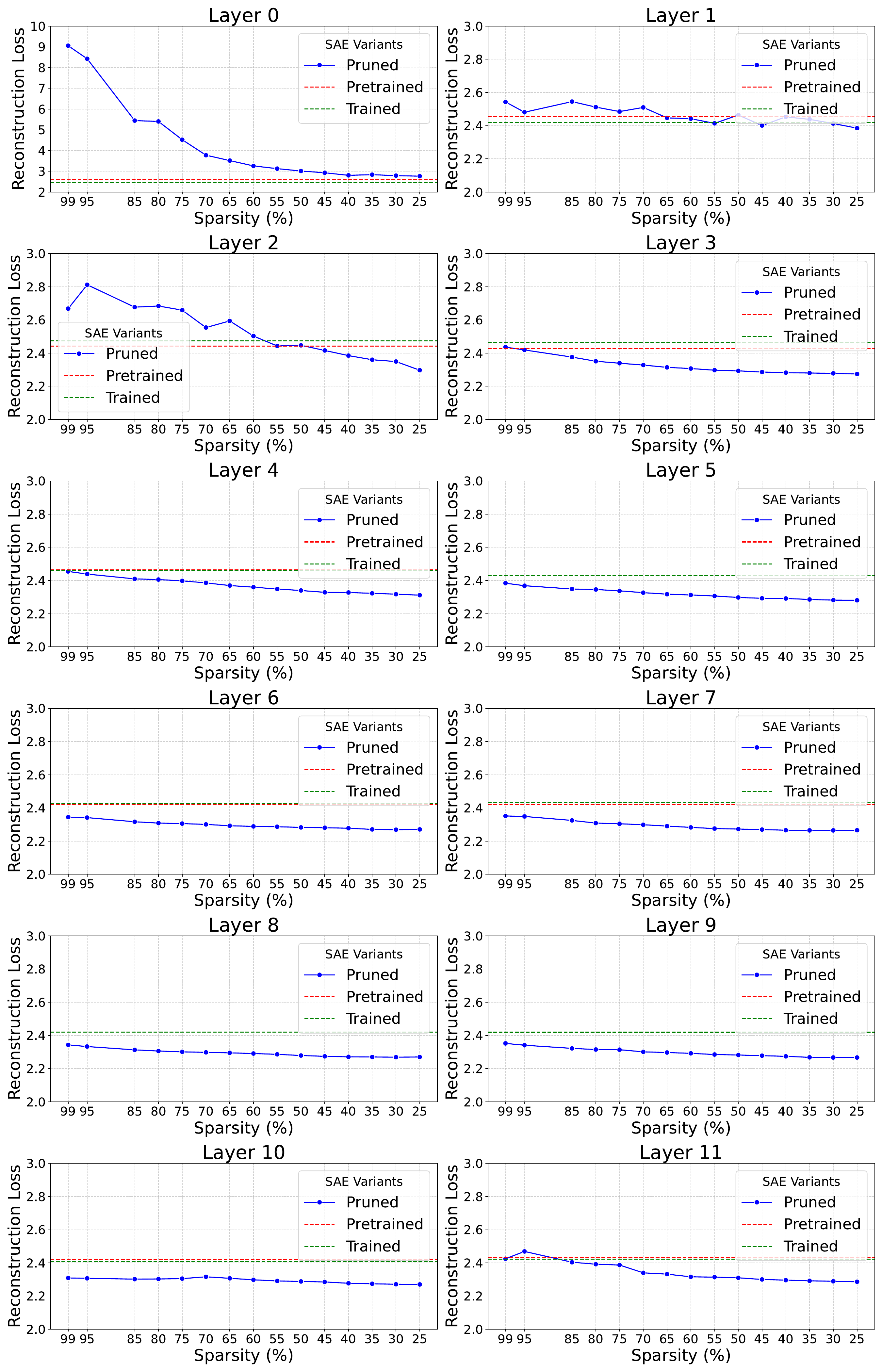}
    \caption{Reconstruction loss for GPT-2 small MLP output}
    \label{fig_appendix: gpt2 mlp recons}
\end{figure*}

\begin{figure*}
    \centering
    \includegraphics[width=\linewidth]{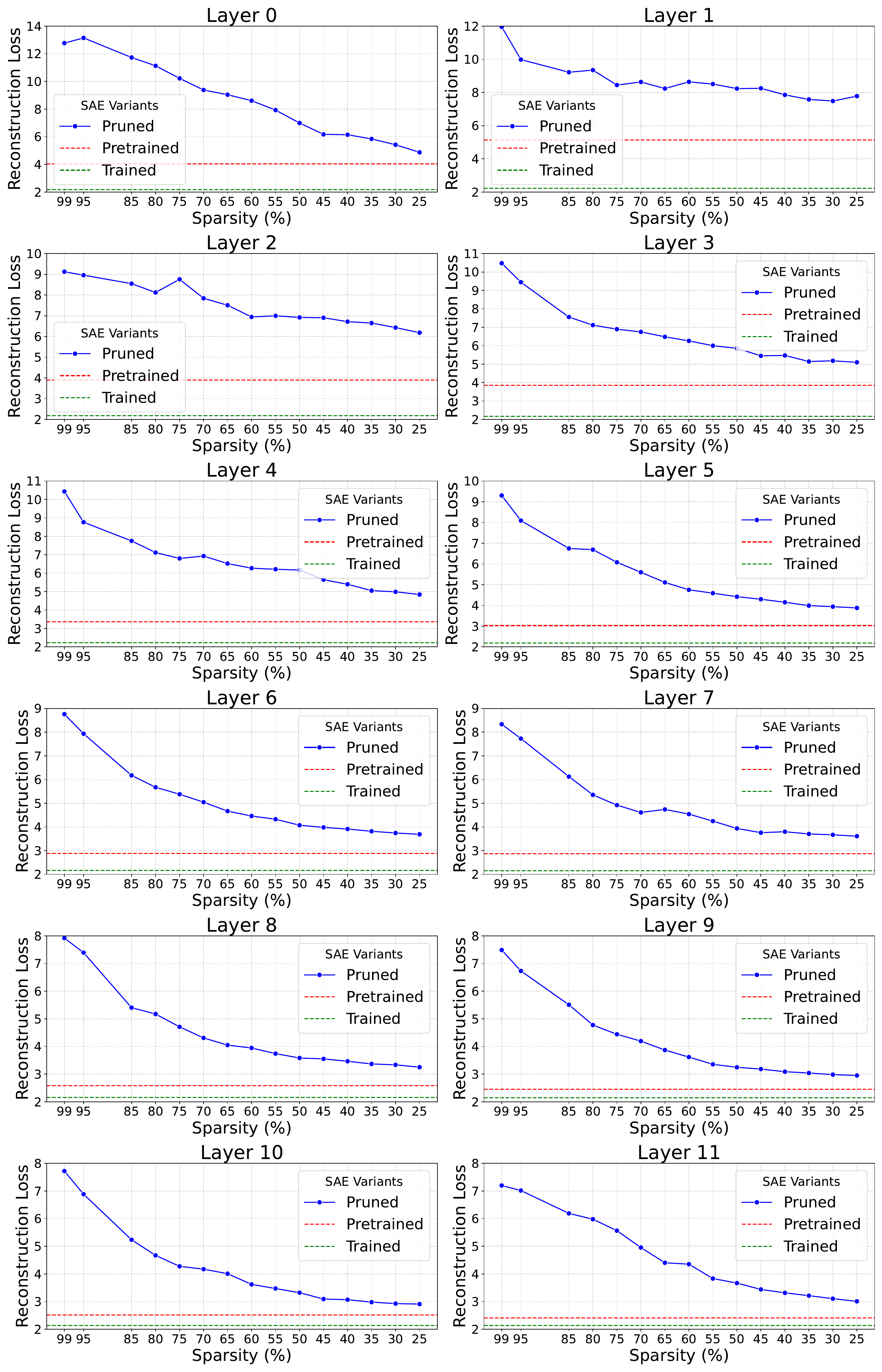}
        \caption{Reconstruction loss for GPT-2 small residual stream output}
    \label{fig_appendix: gpt2 resid recons}
\end{figure*}

\begin{figure*}
    \centering
    \includegraphics[width=\linewidth]{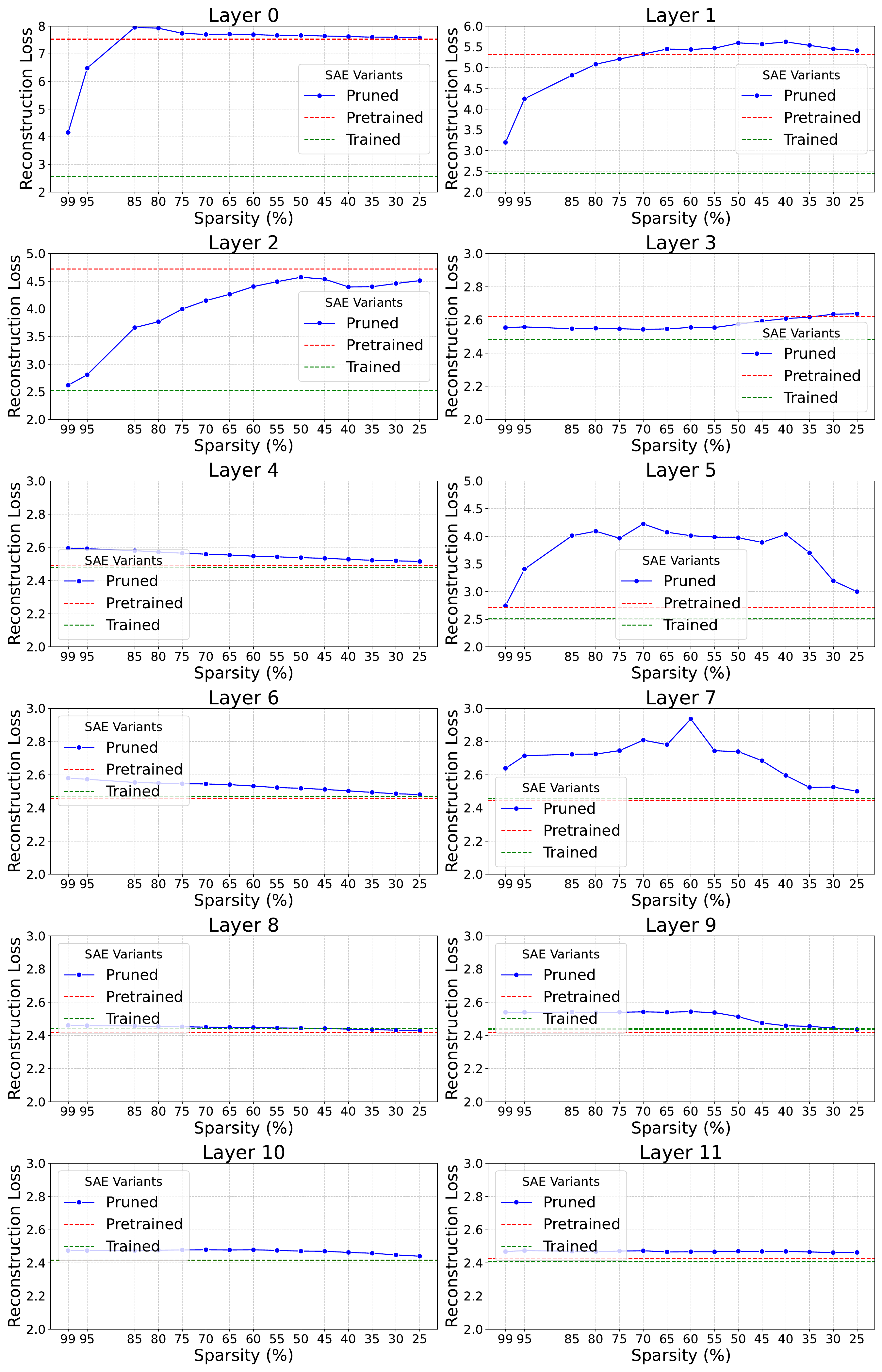}
    \caption{Reconstruction loss for GPT-2 small attention output}
    \label{fig_appendix: gpt2 attn recons}
\end{figure*}

While pruning pretrained Sparse Autoencoders (SAEs) in GPT-2 small, we observed that even at 99\% sparsity, certain layers maintained strong performance. In Figures \ref{fig_appendix: gpt2 mlp recons}, \ref{fig_appendix: gpt2 resid recons}, and \ref{fig_appendix: gpt2 attn recons}, we present layer-wise sparsity ratio tuning results across three outputs.

\subsection{Feature Absorption}\label{A2}
Feature absorption metrics include the following:\\
\textbf{Mean Absorption Fraction Score}:  Represents the average proportion of each feature’s contribution that gets absorbed across all letters. A higher value suggests more consistent partial absorption.\\
\textbf{Mean Full Absorption Score}: Indicates the average degree to which features are fully absorbed (i.e., absorbed with complete contribution) across all letters. \\
\textbf{Mean Number of Split Features}: Reflects the average number of features that were divided or split across different components during absorption.\\
\textbf{Standard Deviation of Absorption Fraction Score}: Measures how much the absorption fraction scores vary from letter to letter.\\
\textbf{Standard Deviation of Full Absorption Score}:  Indicates the variability in the full absorption behavior across letters.\\
\textbf{Standard Deviation of Number of Split Features}: Quantifies how consistently features are being split across the different letters.\\
In Tables \ref{tab_appendix:mlp absorption} and \ref{tab_appendix:resid absorption}, we provide feature absorption scores for MLP and Residual outputs. 

In Tables \ref{tab_appendix:mlp absorption} and \ref{tab_appendix:resid absorption}, we present the absorption scores for MLP and Residual outputs for layer 12 of Gemma-2-2b.
\begin{table*}
\caption{Feature absorption scores for MLP output}
\label{tab_appendix:mlp absorption}
{\footnotesize \begin{tabular}{l|rrrr}
\toprule
\textbf{Metric} & \textbf{Pretrained} & \textbf{Pruned25} & \textbf{Pruned50} & \textbf{Trained} \\
\midrule
Mean Absorption Fraction Score & 0.345 & 0.238 & 0.208 & 0.409 \\
Mean Full Absorption Score     & 0.041 & 0.038 & 0.024 & 0.001 \\
Mean Number of Split Features  & 1.269 & 1.192 & 1.192 & 1.154 \\
Std Dev Absorption Fraction Score & 0.328 & 0.291 & 0.281 & 0.354 \\
Std Dev Full Absorption Score  & 0.065 & 0.055 & 0.026 & 0.002 \\
Std Dev Number of Split Features & 0.533 & 0.491 & 0.634 & 0.368 \\
\bottomrule
\end{tabular}}
\end{table*}

\begin{table*}
\caption{Feature absorption scores for Residual stream output}
\label{tab_appendix:resid absorption}
{\footnotesize \begin{tabular}{l|rrrr}
\toprule
\textbf{Metric} & \textbf{Pretrained} & \textbf{Pruned25} & \textbf{Pruned50} & \textbf{Trained} \\
\midrule
Mean Absorption Fraction Score      & 0.106 & 0.106 & 0.111 & 0.355 \\
Mean Full Absorption Score          & 0.102 & 0.096 & 0.098 & 0.001 \\
Mean Number of Split Features       & 1.154 & 1.269 & 1.154 & 2.385 \\
Std Dev Absorption Fraction Score   & 0.103 & 0.105 & 0.111 & 0.229 \\
Std Dev Full Absorption Score       & 0.102 & 0.103 & 0.111 & 0.003 \\
Std Dev Number of Split Features    & 0.368 & 0.533 & 0.368 & 1.169 \\
\bottomrule
\end{tabular}}
\end{table*}

\subsection{Core (Unsupervised) metrics}\label{A3}
\textbf{Model Behavior Preservation Metrics}: Metrics related to how well the SAE preserves model behavior. These metrics include:
\begin{itemize}[leftmargin=0.75em, labelsep=0.05em, itemsep=0.05em, topsep=0.025em]
    \item \textbf{kl div score} Normalized KL divergence score comparing model behavior with and without SAE.
    \item\textbf{kl div with ablation}: KL divergence when the activation is ablated. 
    \item\textbf{kl div with sae}: KL divergence when using the SAE reconstruction.
\end{itemize}

\textbf{Model Performance Preservation Metrics}: Metrics related to how well the SAE preserves model performance. These metrics include:
\begin{itemize}[leftmargin=1em, labelsep=0.05em, itemsep=0.05em, topsep=0.05em]
    \item\textbf{ce loss score}: Normalized cross entropy loss score comparing model performance with and without SAE 
    \item\textbf{ce loss with ablation}: Cross entropy loss when the activation is ablated.
    \item\textbf{ce loss with sae}: Cross entropy loss when using the SAE reconstruction.
    \item\textbf{ce loss without sae}: Base cross entropy loss without any intervention.
\end{itemize}

\textbf{Reconstruction Quality Metrics}: Metrics related to how well the SAE reconstructs the original activation. These metrics include:
\begin{itemize}[leftmargin=1em, labelsep=0.05em, itemsep=0.05em, topsep=0.05em]
    \item\textbf{explained variance}: Proportion of variance in the original activation explained by the SAE reconstruction 
    \item\textbf{explained variance (legacy)}: Previously used, incorrect, formula for explained variance 
    \item\textbf{mse}: Mean squared error between original activation and SAE reconstruction 
    \item\textbf{cossim}: Cosine similarity between original activation and SAE reconstruction 
\end{itemize}

\textbf{Shrinkage Metrics}: Metrics related to how the SAE changes activation magnitudes.These metrics include:
\begin{itemize}[leftmargin=1em, labelsep=0.05em, itemsep=0.05em, topsep=0.05em]    
    \item\textbf{l2 norm in}: Average L2 norm of input activations 
    \item\textbf{l2 norm out}: Average L2 norm of reconstructed activations 
    \item\textbf{l2 ratio}: Ratio of output to input L2 norms 
    \item\textbf{relative reconstruction bias}: Measure of systematic bias in the reconstruction
\end{itemize}

\textbf{Sparsity Metrics}: Metrics related to feature activation sparsity. These metrics include:
\begin{itemize}[leftmargin=2em, labelsep=0.05em, itemsep=0.25em, topsep=1pt]
\item\textbf{l0 Sparsity}: Average number of non-zero feature activations 
\item\textbf{l1 Sparsity}: Average sum of absolute feature activations 
\end{itemize}

\textbf{Miscellaneous Metrics}: Miscellaneous metrics. These metrics include: 
\begin{itemize}[leftmargin=1em, labelsep=0.05em, itemsep=0.05em, topsep=0.05em]
    \item\textbf{Activation Frequency Over 1\%}: Proportion of tokens that activate each feature more than 1\% of the time 
    \item\textbf{Activation Frequency Over 10\%}: Proportion of tokens that activate each feature more than 10\% of the time 
    \item\textbf{Normalized Activation Frequency Over 1\%}: Sum of $>$ 1\% activation frequency probabilities, normalized by the sum of all feature probabilities 
    \item\textbf{Normalized Activation Frequency Over 10\%}: Sum of $>$ 10\% activation frequency probabilities, normalized by the sum of all feature probabilities 
    \item\textbf{Average Max Encoder Cosine Similarity}: Average of the maximum cosine similarity with any other feature's encoder weights 
    \item\textbf{Average Max Decoder Cosine Similarity}: Average of the maximum cosine similarity with any other feature's decoder weights 
    \item\textbf{Fraction of Alive Features}: Fraction of features that fired at least once during evaluation. This will likely be an underestimation due to a limited amount of tokens \\
\end{itemize}
In Tables \ref{tab_appendix: attn core}, \ref{tab_appendix: mlp core} and \ref{tab_appendix: resid core}, we present the Core metrics for MLP, Attention, and Residual outputs for layer 12 of Gemma-2-2b.

\begin{table*}
\centering
\caption{Core metrics for Attention output }
\label{tab_appendix: attn core}
{\footnotesize \begin{tabular}{l|rrrr}
\toprule
metric & pruned25 & pruned50 & trained & pretrained \\
\midrule
kl div score & 0.821 & 0.775 & 0.824 & 0.852 \\
kl div with ablation & 0.044 & 0.044 & 0.044 & 0.042 \\
kl div with sae & 0.008 & 0.010 & 0.008 & 0.006 \\
\midrule
ce loss score & 1.000 & 0.667 & 1.000 & 0.862 \\
ce loss with ablation & 3.172 & 3.172 & 3.172 & 3.166 \\
ce loss with sae & 3.125 & 3.141 & 3.125 & 3.129 \\
ce loss without sae & 3.125 & 3.125 & 3.125 & 3.124 \\
\midrule
explained variance & 0.793 & 0.727 & 0.859 & 0.802 \\
explained variance legacy & 0.699 & 0.609 & 0.797 & 0.708 \\
mse & 0.037 & 0.049 & 0.025 & 0.035 \\
cossim & 0.891 & 0.855 & 0.930 & 0.893 \\
\midrule
l2 norm in & 19.000 & 19.000 & 19.000 & 18.945 \\
l2 norm out & 17.375 & 16.375 & 16.625 & 17.148 \\
l2 ratio & 0.914 & 0.855 & 0.875 & 0.903 \\
relative reconstruction bias & 1.031 & 1.023 & 0.945 & 1.014 \\
\midrule
l0 & 95.782 & 88.855 & 2835.246 & 88.977 \\
l1 & 103.500 & 94.000 & 189.000 & 98.862 \\
\midrule
freq over 1 percent & 0.114 & 0.099 & 0.708 & 0.091 \\
freq over 10 percent & 0.009 & 0.008 & 0.532 & 0.010 \\
normalized freq over 1 percent & 0.727 & 0.731 & 0.996 & 0.682 \\
normalized freq over 10 percent & 0.348 & 0.339 & 0.954 & 0.378 \\
average max encoder cosine sim & 0.172 & 0.179 & 0.409 & 0.171 \\
average max decoder cosine sim & 0.288 & 0.285 & 0.355 & 0.288 \\
frac alive & 0.963 & 0.960 & 1.000 & 0.963 \\
\bottomrule
\end{tabular}}
\end{table*}

\begin{table*}
\centering
\caption{Core metrics for MLP output }
\label{tab_appendix: mlp core}
{\footnotesize \begin{tabular}{l|rrrr}
\toprule
metric & pruned50 & trained & pretrained & pruned25 \\
\midrule
kl div score & 0.604 & 0.502 & 0.777 & 0.714 \\
kl div with ablation & 0.053 & 0.053 & 0.191 & 0.053 \\
kl div with sae & 0.021 & 0.026 & 0.043 & 0.015 \\
\midrule
ce loss score & 0.667 & 0.667 & 0.848 & 0.667 \\
ce loss with ablation & 3.172 & 3.172 & 5.248 & 3.172 \\
ce loss with sae & 3.141 & 3.141 & 5.085 & 3.141 \\
ce loss without sae & 3.125 & 3.125 & 5.056 & 3.125 \\
\midrule
explained variance & 0.539 & 0.750 & 0.631 & 0.633 \\
explained variance legacy & 0.447 & 0.699 & 0.577 & 0.559 \\
mse & 0.645 & 0.352 & 0.546 & 0.508 \\
cossim & 0.730 & 0.867 & 0.787 & 0.789 \\
\midrule
l2 norm in & 52.750 & 52.750 & 54.276 & 52.750 \\
l2 norm out & 34.500 & 44.000 & 42.296 & 40.000 \\
l2 ratio & 0.656 & 0.832 & 0.776 & 0.758 \\
relative reconstruction bias & 0.902 & 0.973 & 0.988 & 0.957 \\
\midrule
l0 & 61.989 & 4514.003 & 127.725 & 90.342 \\
l1 & 170.000 & 720.000 & 334.119 & 243.000 \\
\midrule
freq over 1 percent & 0.041 & 0.998 & 0.126 & 0.071 \\
freq over 10 percent & 0.007 & 0.902 & 0.013 & 0.008 \\
normalized freq over 1 percent & 0.619 & 1.000 & 0.605 & 0.540 \\
normalized freq over 10 percent & 0.308 & 0.981 & 0.268 & 0.237 \\
average max encoder cosine sim & 0.149 & 0.184 & 0.162 & 0.158 \\
average max decoder cosine sim & 0.158 & 0.220 & 0.163 & 0.163 \\
frac alive & 0.994 & 1.000 & 0.994 & 0.994 \\
\bottomrule
\end{tabular}}
\end{table*}

\begin{table*}
\centering
\caption{Core metrics for Residual stream output }
\label{tab_appendix: resid core}
{\footnotesize \begin{tabular}{l|rrrr}
\toprule
metric & pruned50 & pretrained & pruned25 & trained \\
\midrule
kl div score & 0.981 & 0.989 & 0.989 & 0.978 \\
kl div with ablation & 10.125 & 10.125 & 10.125 & 10.125 \\
kl div with sae & 0.189 & 0.114 & 0.112 & 0.219 \\
\midrule
ce loss score & 0.982 & 0.988 & 0.990 & 0.982 \\
ce loss with ablation & 12.438 & 12.438 & 12.438 & 12.438 \\
ce loss with sae & 3.297 & 3.234 & 3.219 & 3.297 \\
ce loss without sae & 3.125 & 3.125 & 3.125 & 3.125 \\
\midrule
explained variance & 0.828 & 0.867 & 0.867 & 0.883 \\
explained variance legacy & 0.664 & 0.734 & 0.730 & 0.773 \\
mse & 2.156 & 1.680 & 1.703 & 1.492 \\
cossim & 0.898 & 0.918 & 0.918 & 0.938 \\
\midrule
l2 norm in & 153.000 & 153.000 & 153.000 & 153.000 \\
l2 norm out & 134.000 & 140.000 & 140.000 & 157.000 \\
l2 ratio & 0.875 & 0.914 & 0.914 & 1.031 \\
relative reconstruction bias & 0.980 & 0.996 & 1.000 & 1.094 \\
\midrule
l0 & 86.739 & 82.355 & 89.412 & 4658.510 \\
l1 & 512.000 & 544.000 & 568.000 & 1824.000 \\
\midrule
freq over 1 percent & 0.145 & 0.133 & 0.155 & 0.887 \\
freq over 10 percent & 0.002 & 0.002 & 0.002 & 0.811 \\
normalized freq over 1 percent & 0.608 & 0.530 & 0.588 & 0.999 \\
normalized freq over 10 percent & 0.093 & 0.088 & 0.081 & 0.989 \\
average max encoder cosine sim & 0.136 & 0.141 & 0.140 & 0.295 \\
average max decoder cosine sim & 0.270 & 0.278 & 0.277 & 0.347 \\
frac alive & 0.999 & 0.999 & 0.999 & 1.000 \\
\bottomrule
\end{tabular}}
\end{table*}

\subsection{Spurious Correlation Removal (SCR)}\label{A4}
Direction 1 (\textbf{dir 1}): Ablating the top k gender latents to increase profession accuracy\\
Direction 1 (\textbf{dir 2}): Ablating the top k profession latents to increase gender accuracy\\
\textbf{Metric}: Selecting dir1 if initial profession accuracy is lower than initial gender accuracy, else dir2 and ablating the top k SAE latents.

In Tables \ref{tab_appendix:mlp scr} and \ref{tab_appendix:resid scr}, we present the SCR scores for MLP and Residual outputs for layer 12 of Gemma-2-2b.

\begin{table*}
\centering
\caption{SCR scores for MLP output}
\label{tab_appendix:mlp scr}
{\footnotesize \begin{tabular}{l|c|rrrr}
\toprule
metric & k & pruned50 & trained & pretrained & pruned25 \\
\midrule
dir1 & 2 & 0.097 & 0.011 & 0.160 & 0.134 \\
metric & 2 & 0.019 &  0.001 & 0.023 & 0.025 \\
dir2 & 2 & 0.015 &  0.001 & 0.035 & 0.035 \\
\midrule
dir1 & 5 & 0.094 & 0.016 & 0.160 & 0.124 \\
metric & 5 & 0.028 &  0.001 & 0.047 & 0.039 \\
dir2 & 5 & 0.033 & 0.001 & 0.045 & 0.042 \\
\midrule
dir1 & 10 & 0.093 & 0.012 & 0.146 & 0.115 \\
metric & 10 & 0.030 &  0.009 & 0.058 & 0.046 \\
dir2 &  10 & 0.036 &  0.002 & 0.070 & 0.064 \\
\midrule
dir1 & 20 & 0.112 & 0.003 & 0.174 & 0.124 \\
metric & 20 & 0.035 &  0.010 & 0.072 & 0.057 \\
dir2 & 20 & 0.055 & 0.002 & 0.082 & 0.082 \\
\midrule
dir1 & 50 & 0.072 & 0.011 & 0.216 & 0.157 \\
metric & 50 & 0.050 &  0.009 & 0.104 & 0.091 \\
dir2 & 50 & 0.077 & 0.008 & 0.147 & 0.129 \\
\midrule
dir1 & 100 & 0.078 & 0.024 & 0.228 & 0.140 \\
metric & 100 & 0.048 &  0.004 & 0.133 & 0.101 \\
dir2 & 100 & 0.098 & 0.017 & 0.206 & 0.183 \\
\midrule
dir1 & 500 & 0.007 & 0.049 & 0.184 & 0.043 \\
metric & 500 & 0.077 & 0.057 & 0.185 & 0.151 \\
dir2 & 500 & 0.109 & 0.133 & 0.203 & 0.204 \\
\bottomrule
\end{tabular}}
\end{table*}

\begin{table*}
\centering
\caption{SCR scores for Residual stream output}
\label{tab_appendix:resid scr}
{\footnotesize \begin{tabular}{l|c|rrrr}
\toprule
metric & k & pruned50 & pretrained & pruned25 & trained \\
\midrule
dir1 & 2 & 0.262 & 0.286 & 0.295 & 0.062 \\
metric & 2 & 0.100 & 0.112 & 0.111 & 0.018 \\
dir2 & 2 & 0.090 & 0.103 & 0.101 & 0.022 \\
\midrule
dir1 & 5 & 0.287 & 0.318 & 0.329 & 0.099 \\
metric & 5 & 0.143 & 0.170 & 0.167 & 0.036 \\
dir2 & 5 & 0.136 & 0.161 & 0.155 & 0.033 \\
\midrule
dir1 & 10 & 0.290 & 0.322 & 0.341 & 0.146 \\
metric & 10 & 0.222 & 0.243 & 0.245 & 0.047 \\
dir2 & 10 & 0.213 & 0.234 & 0.234 & 0.041 \\
\midrule
dir1 & 20 & 0.219 & 0.287 & 0.299 & 0.115 \\
metric & 20 & 0.269 & 0.304 & 0.306 & 0.076 \\
dir2 & 20 & 0.267 & 0.293 & 0.295 & 0.069 \\
\midrule
dir1 & 50 & 0.166 & 0.278 & 0.300 & 0.153 \\
metric & 50 & 0.340 & 0.365 & 0.368 & 0.101 \\
dir2 & 50 & 0.342 & 0.362 & 0.356 & 0.089 \\
\midrule
dir1 & 100 & 0.146 & 0.210 & 0.222 & 0.148 \\
metric & 100 & 0.306 & 0.401 & 0.316 & 0.128 \\
dir2 & 100 & 0.319 & 0.406 & 0.315 & 0.118 \\
\midrule
dir1 & 500 & 0.132 & 0.126 & 0.157 & 0.035 \\
metric & 500 & 0.304 & 0.349 & 0.338 & 0.203 \\
dir2 & 500 & 0.316 & 0.343 & 0.330 & 0.186 \\
\bottomrule
\end{tabular}}
\end{table*}
\subsection{Targeted Probe Perturbation (TPP)}\label{A5}
Intended: TPP decrease to the intended class only when ablating the top k SAE latents\\
Unintended: TPP decrease to all unintended classes when ablating the top k SAE latents\\
Total metric: TPP metric when ablating the top k SAE latents.

In Tables \ref{tab_appendix:mlp tpp} and \ref{tab_appendix:resid tpp}, we present the TPP scores for MLP and Residual outputs for layer 12 of Gemma-2-2b.
\begin{table*}
\centering
\caption{TPP scores: MLP output}
\label{tab_appendix:mlp tpp}
{\footnotesize \begin{tabular}{l|c|rrrr}
\toprule
metric & k & pruned50 & trained & pretrained & pruned25 \\
\midrule
Total & 2 & 0.004 & 0.003 & 0.004 & 0.003 \\
Intended  & 2 & 0.005 & 0.004 & 0.006 & 0.005 \\
 Unintended  & 2 & 0.001 & 0.001 & 0.002 & 0.002 \\
\midrule
 Total & 5 & 0.010 & 0.004 & 0.012 & 0.012 \\
 Intended  & 5 & 0.012 & 0.005 & 0.013 & 0.013 \\
 Unintended  & 5 & 0.002 & 0.001 & 0.002 & 0.002 \\
\midrule
 Total & 10  & 0.014 & 0.004 & 0.015 & 0.015 \\
 Intended   & 10 & 0.016 & 0.005 & 0.018 & 0.018 \\
 Unintended  & 10 & 0.002 & 0.001 & 0.003 & 0.003 \\
\midrule
Total & 20 & 0.026 & 0.007 & 0.029 & 0.029 \\
Intended  & 20 & 0.031 & 0.009 & 0.034 & 0.034 \\
Unintended  & 20 & 0.004 & 0.002 & 0.005 & 0.004 \\
\midrule
Total & 50 & 0.063 & 0.023 & 0.090 & 0.076 \\
Intended  & 50 & 0.069 & 0.027 & 0.098 & 0.084 \\
Unintended  & 50 & 0.006 & 0.004 & 0.009 & 0.008 \\
\midrule
Total & 100 & 0.115 & 0.069 & 0.168 & 0.145 \\
Intended  & 100 & 0.124 & 0.074 & 0.181 & 0.156 \\
Unintended  & 100 & 0.008 & 0.005 & 0.013 & 0.011 \\
\midrule
Total & 500 & 0.217 & 0.303 & 0.351 & 0.316 \\
Intended  & 500 & 0.236 & 0.343 & 0.383 & 0.342 \\
Unintended  & 500 & 0.019 & 0.039 & 0.032 & 0.026 \\
\bottomrule
\end{tabular}}
\end{table*}

\begin{table*}
\centering
\caption{TPP scores: Residual stream output}
\label{tab_appendix:resid tpp}
{\footnotesize \begin{tabular}{l|c|rrrr}
\toprule
metric  & k & pruned50 & pretrained & pruned25 & trained \\
\midrule
Total & 2 & 0.007 & 0.008 & 0.008 & 0.008 \\
intended  & 2 & 0.010 & 0.011 & 0.011 & 0.010 \\
Unintended  & 2 & 0.003 & 0.003 & 0.002 & 0.002 \\
\midrule
Total & 5 & 0.012 & 0.015 & 0.014 & 0.011 \\
Intended  & 5 & 0.014 & 0.018 & 0.016 & 0.014 \\
 Unintended  & 5 & 0.003 & 0.003 & 0.003 & 0.002 \\
\midrule
 Total & 10 & 0.018 & 0.022 & 0.021 & 0.012 \\
Intended  & 10 & 0.022 & 0.026 & 0.025 & 0.019 \\
 Unintended  & 10  & 0.004 & 0.004 & 0.004 & 0.006 \\
\midrule
 Total & 20 & 0.041 & 0.046 & 0.047 & 0.028 \\
Intended  & 20 & 0.047 & 0.053 & 0.054 & 0.038 \\
 Unintended  & 20 & 0.007 & 0.007 & 0.007 & 0.010 \\
\midrule
 Total & 50  & 0.090 & 0.112 & 0.116 & 0.084 \\
Intended  & 50  & 0.097 & 0.121 & 0.124 & 0.097 \\
 Unintended  & 50  & 0.008 & 0.009 & 0.008 & 0.013 \\
\midrule
 Total & 100 & 0.157 & 0.201 & 0.213 & 0.171 \\
Intended  & 100 & 0.167 & 0.211 & 0.224 & 0.190 \\
 Unintended  & 100 & 0.010 & 0.010 & 0.010 & 0.019 \\
\midrule
 Total & 500 &  0.370 & 0.395 & 0.402 & 0.353 \\
Intended  & 500 & 0.387 & 0.412 & 0.420 & 0.423 \\
 Unintended  & 500 & 0.017 & 0.016 & 0.018 & 0.070 \\
\bottomrule
\end{tabular}}
\end{table*}

\subsection{RAVEL}\label{A6}
Disentanglement Score: Mean of cause and isolation scores across RAVEL datasets.\\
Cause Score: Patching attribute-related SAE latents. High cause accuracy indicates that the SAE latents are related to the attribute.
Isolation score: Patching SAE latents related to another attribute. High isolation accuracy indicates that latents related to another attribute are not related to this attribute.\\
To establish entity attribute pairs, City: [Country, Continent, Language] and Nobel Prize Winners: [Birth Country, Field, Gender].  

In Table \ref{tab_appendix: RAVEL}, we present the RAVEL scores for MLP and Residual outputs for layer 12 of Gemma-2-2b.

\begin{table*}
\centering
\caption{RAVEL metrics for different entity-relation pairs across SAE variants.}
\label{tab_appendix: RAVEL}
{\footnotesize 
\begin{tabular}{l|l|l|cccc}
\toprule
\textbf{Out} & \textbf{Pairs} & \textbf{Score} & \textbf{Pretrained} & \textbf{Pruned25} & \textbf{Pruned50} & \textbf{Trained} \\
\midrule
\multirow{18}{*}{Resid} & \multirow{3}{*}{1}  & Cause     & 0.7051 & 0.6585 & 0.7690 & 0.8003 \\
& & Isolation     & 0.4599 & 0.4673 & 0.4379 & 0.4833 \\
& & Disentangle   & 0.5825 & 0.5629 & 0.6034 & 0.6418 \\
\cline{2-7}\addlinespace
& \multirow{3}{*}{2} & Cause   & 0.3895 & 0.4129 & 0.4857 & 0.5884 \\
& & Isolation     & 0.8119 & 0.8228 & 0.7421 & 0.7558 \\
& & Disentangle   & 0.6007 & 0.6179 & 0.6139 & 0.6721 \\
\cline{2-7} \addlinespace
& \multirow{3}{*}{3}  & Cause         & 0.5971 & 0.6004 & 0.6258 & 0.6118 \\
& & Isolation     & 0.5359 & 0.5259 & 0.4874 & 0.5445 \\
& & Disentangle   & 0.5665 & 0.5631 & 0.5566 & 0.5781 \\
\cline{2-7} \addlinespace
& \multirow{3}{*}{4}  & Cause         & 0.6521 & 0.6683 & 0.5745 & 0.4255 \\
& & Isolation     & 0.9324 & 0.9400 & 0.9135 & 0.8377 \\
& & Disentangle   & 0.7922 & 0.8042 & 0.7440 & 0.6316 \\
\cline{2-7}\addlinespace
& \multirow{3}{*}{5}  & Cause         & 0.7989 & 0.8056 & 0.7707 & 0.7411 \\
& & Isolation     & 0.9471 & 0.9392 & 0.9352 & 0.8526 \\
& & Disentangle   & 0.8730 & 0.8724 & 0.8530 & 0.7969 \\
\cline{2-7}\addlinespace
& \multirow{3}{*}{6}  & Cause         & 0.6084 & 0.6503 & 0.5221 & 0.5431 \\
& & Isolation     & 0.9486 & 0.9443 & 0.9251 & 0.8865 \\
& & Disentangle   & 0.7785 & 0.7973 & 0.7236 & 0.7148 \\
\cline{1-7}\addlinespace
\multirow{18}{*}{MLP} &  \multirow{3}{*}{1}  & Cause         & 0.7270 & 0.5985 & 0.3322 & 0.0652 \\
& & Isolation     & 0.3632 & 0.3391 & 0.3298 & 0.3244 \\
& &  Disentangle   & 0.5451 & 0.4688 & 0.3310 & 0.1948 \\
\cline{2-7}\addlinespace
&  \multirow{3}{*}{2} & Cause         & 0.5585 & 0.5299 & 0.4174 & 0.2055 \\
& & Isolation     & 0.6259 & 0.5431 & 0.4234 & 0.2469 \\
& & Disentangle   & 0.5922 & 0.5365 & 0.4204 & 0.2262 \\
\cline{2-7}\addlinespace
&  \multirow{3}{*}{3}  & Cause         & 0.4685 & 0.4090 & 0.2892 & 0.0964 \\
& & Isolation     & 0.4807 & 0.4396 & 0.3951 & 0.3792 \\
& & Disentangle   & 0.4746 & 0.4243 & 0.3421 & 0.2378 \\
\cline{2-7}\addlinespace
& \multirow{3}{*}{4}  & Cause         & 0.2816 & 0.2089 & 0.1150 & 0.0833 \\
& & Isolation     & 0.7846 & 0.7757 & 0.7644 & 0.7018 \\
& & Disentangle   & 0.5331 & 0.4923 & 0.4397 & 0.3926 \\
\cline{2-7}\addlinespace
& \multirow{3}{*}{5} & Cause         & 0.5797 & 0.5367 & 0.4512 & 0.2905 \\
& & Isolation     & 0.8321 & 0.8229 & 0.8083 & 0.7865 \\
& & Disentangle   & 0.7059 & 0.6798 & 0.6298 & 0.5385 \\
\cline{2-7}\addlinespace
&\multirow{3}{*}{6} & Cause         & 0.8625 & 0.8625 & 0.7972 & 0.4289 \\
& & Isolation     & 0.8051 & 0.7880 & 0.7366 & 0.6574 \\
& & Disentangle   & 0.8338 & 0.8252 & 0.7669 & 0.5431 \\
\bottomrule
\end{tabular}}
\end{table*}

\subsection{Sparse Probing}\label{A7}
Sparse probing builds on \cite{gurnee2023findingneuronshaystackcase}, which used sparse probing to find MLP neurons tied to specific contexts, and \cite{gao2024scaling} extended the idea to SAEs. This method tests whether an SAE captures specific, predefined concepts in distinct latents. For a given concept, the method identifies the top k latent dimensions that respond most differently to positive versus negative examples. Then, a probe is trained using only those top k latents. If these latents truly represent the concept, the probe should perform well, even though the SAE was never explicitly trained to separate that concept. This approach is efficient because it relies only on the SAE and the precomputed model activations.

\begin{table*}
\centering
\caption{Sparse probing across three output types and four SAE variants}
\label{tab_appendix: sparse probing}
{\footnotesize \begin{tabular}{l|l|rrrr}
\toprule
Output & Test Accuracy & Pretrained & Pruned25 & Pruned50 & Trained \\
\midrule
\multirow{4}{*}{Residual}  & Overall & 0.954 & 0.955 & 0.956 & 0.956 \\
& Top 1 & 0.768 & 0.768 & 0.762 & 0.748 \\
& Top 2 & 0.802 & 0.804 & 0.804 & 0.799 \\
& Top 5 & 0.869 & 0.871 & 0.869 & 0.851 \\
\midrule
\multirow{4}{*}{MLP}  & Overall & 0.944 & 0.943 & 0.937 & 0.942 \\
& Top 1 & 0.661 & 0.712 & 0.703 & 0.712 \\
& Top 2 & 0.691 & 0.765 & 0.759 & 0.766 \\
& Top 5 & 0.750 & 0.835 & 0.822 & 0.829 \\
\bottomrule
\end{tabular}}
\end{table*}

\end{document}